%% file: main.tex
\documentclass{article}

\usepackage{PRIMEarxiv}

\usepackage{lineno,hyperref}
\usepackage{amsmath}
\usepackage{amsfonts}
\usepackage{amssymb}
\usepackage{geometry}
\usepackage{txfonts}
\usepackage{graphics}
\usepackage{graphicx}
\usepackage{bm}
\usepackage{booktabs}
\usepackage{multirow}
\usepackage{threeparttable}
\usepackage{caption}
\usepackage{subcaption}
\usepackage[utf8]{inputenc} 
\usepackage[T1]{fontenc}    
\usepackage{url}            
\usepackage{nicefrac}       
\usepackage{microtype}      
\usepackage{fancyhdr}       
\usepackage{tikz}
\usepackage{lipsum}
\usepackage{cancel}
\usepackage{bm}         
\usepackage{mathtools, nccmath}
\usepackage{xspace}
\usepackage{bbm}
\usepackage{hyperref}
\usepackage[capitalize]{cleveref}
\usepackage{siunitx} 
\usepackage[normalem]{ulem}
\usepackage{comment}
\usepackage{booktabs}	
\usepackage{float}	
\usepackage{booktabs} 
\usepackage{adjustbox} 
\usepackage{array} 
\usepackage{roboto} 
\usepackage{siunitx} 
\usepackage{threeparttable}
\usepackage{gensymb}
\usepackage{algorithm}
\usepackage{algpseudocode}
\usepackage{csquotes}
\usetikzlibrary{matrix, positioning, arrows}
\usepackage{rotating}

\usetikzlibrary{arrows}
\usepackage{caption}

\usepackage{tikz, pgfplots}
\usetikzlibrary{positioning,shadows}
\pgfplotsset{compat=1.18}
\definecolor{light_blue}{HTML}{ADDFFF}
\definecolor{purple}{HTML}{9400d3}
\usetikzlibrary{calc} 

\hypersetup{
    colorlinks=true,      
    linkcolor=black,      
    citecolor=black,      
    filecolor=black,      
    urlcolor=black        
}

\newcommand{\brc}[1]{\left(#1\right)}       
\newcommand{\cbk}[1]{\left\{#1\right\}}     
\newcommand{\sbk}[1]{\left[#1\right]}       

\newcommand{\norm}[1]{\left\lVert#1\right\rVert}

\newcommand{\ve}[1]{\bm{{#1}}}

\usepackage[numbers]{natbib}
\usepackage{setspace}

\title{A RECURSIVE BAYESIAN NEURAL NETWORK FOR CONSTITUTIVE MODELING OF SANDS UNDER MONOTONIC AND CYCLIC LOADING.}

\title{A recursive Bayesian neural network for constitutive modeling of sands under monotonic and cyclic loading}

\author{Toiba Noor\\
  Department of Civil Engineering\\
  Indian Institute of Technology Delhi\\
  \texttt{cez217542@iitd.ac.in} \\
  \And
      Soban Nasir Lone \\
  Department of Transportation Systems Engineering\\
  Technical University of Munich\\
  \texttt{soban.lone@tum.de} \\
  \And
      G. V. Ramana \\
  Department of Civil Engineering\\
  Indian Institute of Technology Delhi\\
  \texttt{ramana@civil.iitd.ac.in} \\
  \And
      Rajdip Nayek \\
  Department of Applied Mechanics\\
  Indian Institute of Technology Delhi\\
  \texttt{rajdipn@am.iitd.ac.in} \\   
}

\usepackage{tabularx}
\usepackage{longtable}
\usepackage{enumitem}
\newcolumntype{C}{>{\centering\arraybackslash}X} 

\begin{document}
\maketitle

\begin{abstract}
In geotechnical engineering, constitutive models are central to capturing soil behavior across diverse drainage conditions, stress paths, and loading histories. While data-driven deep learning approaches have shown promise as alternatives to traditional constitutive formulations, their practical deployment requires models that are not only accurate but also capable of quantifying predictive uncertainty. This study introduces a recursive Bayesian neural network (rBNN) framework that unifies temporal sequence learning with generalized Bayesian inference to achieve both predictive accuracy and principled uncertainty quantification. A key innovation is a sliding-window recursive structure that captures path-dependent soil responses under monotonic and cyclic loading. Each model is dataset-specific and predicts the response of a triaxial specimen to a prescribed strain history from its measured initial state. By treating network parameters as random variables and inferring their posterior distributions via generalized variational inference, the rBNN produces confidence intervals alongside point predictions, whose empirical coverage of the observed responses is quantified on held-out tests.

The framework is validated against four datasets spanning both simulated and experimental triaxial tests: monotonic loading using a Hardening Soil model simulation and 28 consolidated drained tests on Baskarp sand, and cyclic loading using an exponential constitutive simulation of drained and undrained tests and 35 experimental cyclic consolidated undrained tests on Ottawa F65 sand. 

All models, including long short-term memory, gated recurrent unit and encoder--decoder networks and two uncertainty baselines (deep ensembles and Monte Carlo dropout), were tuned with an equal search budget and evaluated on held-out tests. The rBNN is nearly as accurate as the best recurrent networks (mean coefficient of determination 0.95--0.99), and its 95\% predictive intervals cover at least 90\% of the measured response on 19 of the 21 test specimens, against 13 of 21 for a deep ensemble of the same recursive architecture, while being narrower than the Monte Carlo dropout intervals on three of the four datasets. For the Ottawa F65 tests, the predicted range of liquefaction onset contains the measured onset of all three test specimens. An ablation study shows that training over recursive windows is what keeps long step-by-step predictions stable, and a comparison of single networks shows that Bayesian weights make the uncertainty bands considerably more reliable than a variance output alone.
\end{abstract}


\section{Introduction}\label{sec:intro}
Constitutive modeling enables prediction of soil response under diverse drainage conditions, stress paths, and loading histories. Unlike metals, soils --- particularly sands --- exhibit nonlinear, path-dependent behavior due to their granular structure, density, and stress history \cite{hong2017small,yin2017modeling}. Accurate models are essential for designing and safety assessment of foundations, tunnels, offshore structures, and landslide-prone slopes \cite{ishihara1996soil,hejazi2008impact,secondi2013landslide,randolph2017offshore}.

Traditional constitutive models --- including elastic, viscoelastic, elastoplastic, and micromechanical formulations \cite{drucker1952soil,makris1993models,katona1984evaluation} --- describe features such as hardening, dilatancy, and pore pressure evolution \cite{yao2019unified,cassiani2017strain,dafalias2021elastoplastic,yin2020elastoplastic}. However, they demand extensive parameter calibration, offer limited generalization, and may suffer from numerical instability \cite{yin2019practice,herle2021fundamentals}. These limitations are acute for sands, which display dilatancy \cite{schanz1996angles}, state dependence \cite{yang2004state}, and anisotropy \cite{wu1998rational}, especially under cyclic loading. 

To overcome these limitations, data-driven approaches, especially deep learning (DL) techniques, have emerged as promising alternatives. They learn stress--strain relationships directly from data, bypassing explicit formulations and extensive parameter calibration. Early applications used feedforward neural networks (FFNNs) \cite{ghaboussi1991knowledge,ellis1992neural,ellis1995stress} to model monotonic loading \cite{sidarta1998constitutive,penumadu1999triaxial,habibagahi2003neural}
(see \cref{fig:fnn}). However, FFNNs are static and accumulate error during sequential prediction. Recursive FFNNs (rFFNNs) address this by incorporating their own predictions into subsequent inputs (see \cref{fig:fbnn})\cite{najjar2000characterizing,ghaboussi1998new}. Typically rFFNNs are trained to predict the \textit{entire series} from the initial stress value \textit{all at once}, aligning the training process with the prediction strategy. While theoretically superior, in practice this often causes error accumulation: small inaccuracies at early steps propagate and amplify, producing phase shifts and amplitude errors, especially in cyclic loading \cite{basheer2000selection}. Although the loss averages over all steps, early errors are hard to correct since gradients diminish or destabilize through many recursions \cite{bengio1994learning,bengio2015scheduled}.

In the current study, we propose a revised training strategy for rFFNNs based on overlapping sliding windows. The full loading sequence is divided into shorter, fixed-length windows, each of which recursively combines its own predictions with inputs, mimicking the inference procedure but over a limited horizon. This constrains recursion depth, improves gradient flow, and allows localized error correction. The window length acts as a tunable hyperparameter that balances temporal context with training stability. Inference remains autoregressive over the full sequence, but training becomes more stable, which improves accuracy especially under cyclic loading.

\input{figure1.tex}

Recurrent neural networks (RNNs), including Long Short-Term Memory (LSTM) and Gated Recurrent Units (GRUs), are the main alternative for sequential modeling. RNNs naturally advanced sand constitutive modeling by maintaining memory over multiple load steps by summarizing previous information through ``hidden states'', thus capturing temporal dependencies beyond first-order Markov models. However, early RNNs struggled with vanishing and exploding gradients, limiting their ability to represent stress--strain--volume change behavior of granular materials \cite{zhuu1998modeling,najjar1999simulating,romo2001recurrent}. To overcome these issues, advanced architectures like LSTM \cite{hochreiter1997long} and GRU \cite{chung2014empirical} introduced gating mechanisms that regulate information flow, allowing longer memory and mitigating gradient problems. These models have proven effective for capturing complex sand constitutive behavior \cite{zhang2021state,zhang2021application,qu2021deep,wu2023constitutive,zhang2022physics}, particularly under cyclic loading where path dependency is critical. Notably, LSTMs and GRUs have demonstrated strong performance in drained and undrained conditions \cite{zhang2020ai}, including tasks such as predicting shear strain, excess pore pressure, and liquefaction potential \cite{jas2024prediction,birinci2025prediction,jas2025evaluation}. Recent comparative studies distinguish \textit{cycle-based} models, which predict cycle-wise quantities such as the number of cycles to liquefaction, from \textit{time history-based} models, which predict the complete response history \cite{jas2025evaluation,jas2026comparative}; the present framework belongs to the latter class.

Encoder--decoder architectures extend recurrent models by separating input encoding from output generation, supporting structured sequence-to-sequence prediction and long-range dependency capture \cite{cho2014learning,du2020multivariate}. Here, the encoder is a bidirectional LSTM (BiLSTM) that incorporates information from past and future load steps of the prescribed strain path (which is known a priori), while the decoder is a unidirectional LSTM generating outputs in a causal manner. To our knowledge, such an encoder--decoder formulation has not been applied to sand constitutive modeling and is introduced here as a deterministic sequence-to-sequence baseline under comparable training and inference conditions.

While ML and DL models offer attractive alternatives in capturing complex relationships without the need for explicit physical laws, they also present challenges, such as optimizing training techniques, managing limited and noisy data, effectively quantifying uncertainties, and minimizing overfitting risks. Simulated datasets are limited in size, and experimental datasets are in addition sparse and noisy, so that predictions carry both limited-data and measurement uncertainty. Deterministic models (FFNNs, rFFNNs, RNNs and encoder--decoders) yield only point estimates, whereas uncertainty quantification (UQ) provides confidence bounds that support better-informed geotechnical decisions.

To overcome these limitations, we enhance the FFNN architecture by incorporating uncertainty modeling, treating the model parameters $\ve{\beta}$ as random variables, and introducing a deterministic uncertainty layer as the final layer in the FFNN. By deterministic uncertainty layer, we mean that the output is a distribution (in our case Gaussian distribution), with the mean and standard deviation as a deterministic function of the FFNN parameters. As such, the model parameters $\ve{\beta}$ becomes amenable to Bayesian inference, akin to Bayesian neural networks (BNNs) \cite{mackay1995probable,neal2012bayesian,jospin2022hands}.

However, as will be explained later in \cref{sec:rBNN}, a recursive loss-based formulation over multiple overlapping sliding windows causes a departure from likelihood-based formulation. Therefore, we adopt a \textit{generalized} Bayesian inference framework, which allows the use of loss function instead of likelihood function \cite{JMLR:v23:19-1047,ghosh2016robust,guedj2019primer}. Additionally, Bayesian approaches mitigate overfitting by automatically regularising the models by using a prior distribution as inductive bias.
BNNs have been applied in soil constitutive modeling for regularization \cite{pham2022bayesian}.
Similarly, methods like Monte Carlo (MC) dropout \cite{gal2016dropout}, have been used for regularization but can also be used for UQ \cite{zhang2023interpretable}. However, MC dropout does not provide the comprehensive probabilistic modeling that a fully Bayesian approach offers \cite{gal2016dropout}. Deep ensembles, in which several networks with a variance output are trained independently and their predictions pooled, are a strong non-Bayesian alternative \cite{lakshminarayanan2017simple}. Both MC dropout and deep ensembles are therefore included as uncertainty baselines in this study. 

To the best of the authors' knowledge, this is the first framework for constitutive modeling of sands that combines a recursive (autoregressive) neural network with generalized variational Bayesian inference, termed here a recursive Bayesian neural network (rBNN). The contributions are: (a) a sliding-window training strategy for rFFNNs, (b) a Bayesian extension with uncertainty-aware predictions, and (c) benchmarking against deterministic sequence models and against deep-ensemble and MC dropout baselines, all tuned with an equal search budget, under monotonic and cyclic loading, including the coverage of the predictive intervals on every test specimen, the predicted onset of liquefaction , and (d) an ablation study of recursive training and Bayesian weights. 
The scope of this study is dataset-specific prediction under prescribed triaxial strain histories: for each dataset, a separate model is trained to predict the stress and volumetric or pore pressure response of a specimen to a prescribed axial strain history, given its initial state and test constants. The models are not general constitutive laws for other sands, stress paths or boundary-value problems. The remainder of this paper describes the problem setup (\cref{sec:probsetup}), methodology (\cref{sec:method}), performance assessment (\cref{sec:assess}), discussion and limitations (\cref{sec:discussion}), and conclusions (\cref{sec:concusion}).

\section{Problem Setup} \label{sec:probsetup}
This study aims to develop uncertainty-aware probabilistic constitutive models for sandy soils under quasi-static triaxial loading. Two separate modeling frameworks are proposed, each trained on data from $M$ strain-controlled tests: one for \textit{monotonic} loading under consolidated drained (CD) conditions, and another for \textit{cyclic} loading under drained and undrained conditions (simulated data) and consolidated undrained (CU) conditions (experimental data). The choice aligns with standard geotechnical practice for sands: CD tests characterize strength and volumetric response under drained conditions, while undrained cyclic tests replicate rapid loading scenarios---such as seismic events---where drainage is restricted and excess pore pressure accumulation governs soil instability and liquefaction. 

\begin{figure}[!ht]
     \centering
     \includegraphics[width=0.45\textwidth]{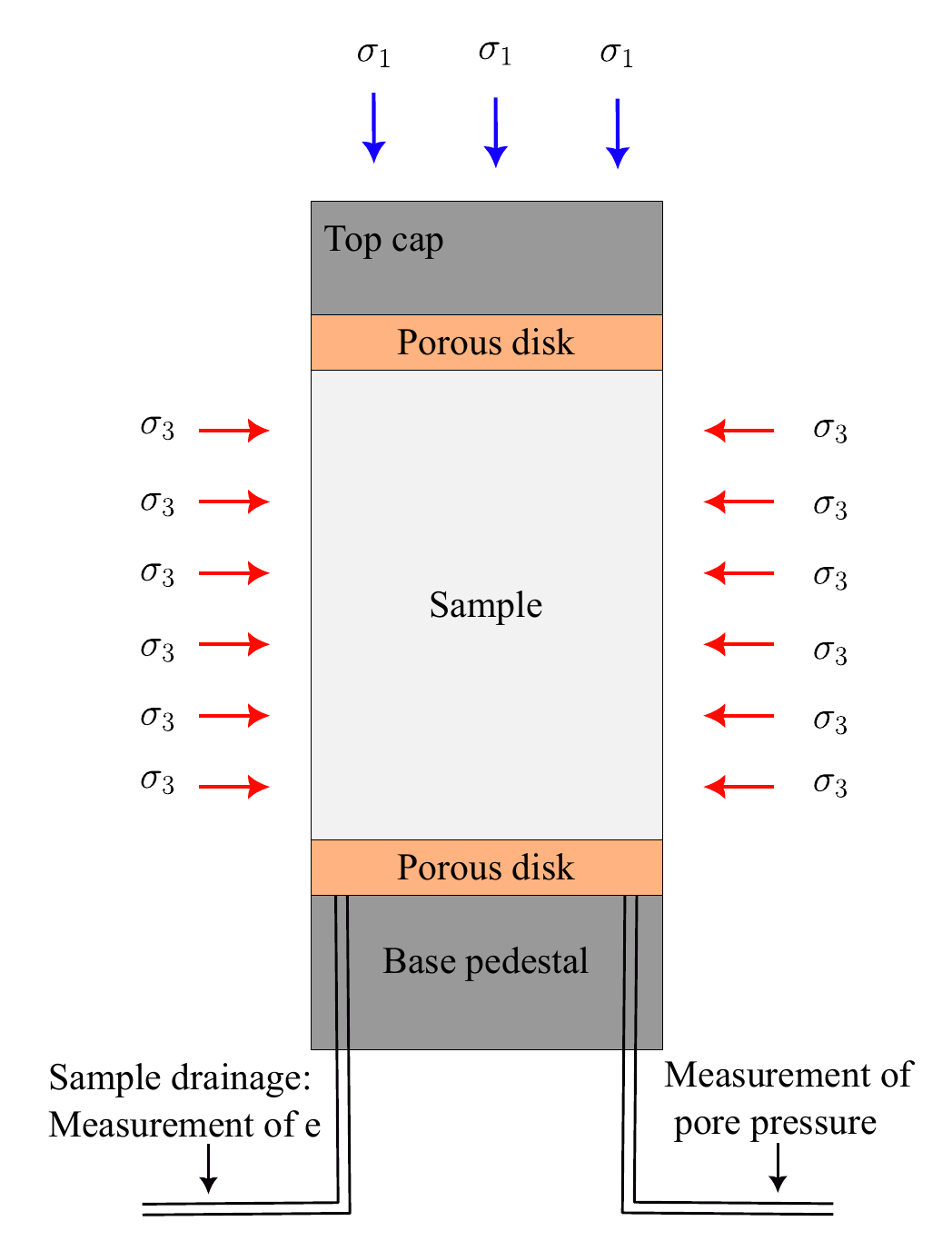}
     \caption{Schematic of the triaxial test setup.}
     \label{fig:cd}
\end{figure}

Each test records the stress-strain-volume (or pore pressure) responses of a sand sample under controlled loading conditions. Initially, the sand sample is consolidated under a constant effective confining pressure \( \sigma'_3 \), and then incremental axial strains \( \text{d}\epsilon \) are applied through quasi-static loading (monotonic compression, or alternating loading and unloading in cyclic tests), illustrated in \cref{fig:cd}. These load steps represent pseudo-time indices, allowing the modeling of sequential soil responses. At each load step, the mean effective stress $(p')$ and deviatoric stress ($q$) are computed as:
\begin{align}
    p' = \frac{\sigma'_1 + 2\sigma'_3}{3}, \quad q = \sigma'_1 - \sigma'_3
\end{align}
where $\sigma'_1$ and $\sigma'_3$ denote the effective axial and radial stresses, respectively.
The interpretation of the third state parameter varies with the type of test. In monotonic CD tests, the void ratio \(e\) is directly computed from the initial void ratio \(e_0\) and the measured volume change, modified due to the rearrangement of particles. In contrast, in the undrained cyclic tests no volume change occurs; instead, the evolution of pore pressure is tracked through the excess pore pressure ratio \(r_u\). For ease of notation, \(e\) denotes the third state variable (\(e\), \(r_u\) or the volumetric strain \(\epsilon_v\)) in the equations, while figures and tables use the physical variables.

Each triaxial test \( m \in \{1, 2, \dots, M\} \) consists of multiple load steps \( t \in \{1, \dots, N\} \), with a state vector at each step defined as:
\begin{align}
\ve{s}^{(m)}_{t} = \begin{bmatrix} {p'}^{(m)}_{t}, \; q^{(m)}_{t}, \; e^{(m)}_{t} \end{bmatrix}^T
\end{align}

It is influenced by a set of exogenous inputs $\ve{u}_t^{(m)}$, which vary depending on the loading type:

\begin{itemize}
    \item \textbf{Monotonic loading:} $\ve{u}_t^{(m)} = \cbk{\epsilon_t^{(m)},\; \text{d}\epsilon_t^{(m)}}$
    \item \textbf{Cyclic loading:} $\ve{u}_t^{(m)} = \cbk{\epsilon_t^{(m)},\; \text{d}\epsilon_t^{(m)},\; \Delta_t^{(m)},\; c_t^{(m)}}$
\end{itemize}
Here, $\epsilon_t^{(m)}$ is the axial strain, $\text{d}\epsilon_t^{(m)}$ is its increment, $\Delta_t^{(m)} \in \{-1, +1\}$ denotes the strain direction (loading or unloading), and $c_t^{(m)}$ is the current cycle number. In the cyclic datasets, $\text{d}\epsilon_t^{(m)}$ is replaced in the implementation by the sign of $\epsilon_t^{(m)}$ (compression or extension side of the cycle), because the strain increments are nearly constant in magnitude and their sign is carried by $\Delta_t^{(m)}$. Both $\Delta_t^{(m)}$ and $c_t^{(m)}$ are derived from the prescribed strain history and carry no information about the stress or pore pressure response. Additionally, each test is associated with triaxial test constants $\ve{\theta}^{(m)} = \cbk{{\sigma'_3}^{(m)},\; e_0^{(m)}}$. In the implementation, $\ve{\theta}^{(m)}$ contains the constants that vary within each dataset: $\log\sigma'_3$, $E_{50}^{\text{ref}}$ and the friction angle $\phi$ (simulated monotonic); $\log\sigma'_3$ and $e_0$ (Baskarp sand); $e_0$ and the drainage condition, which fixes the initial mean stress (simulated cyclic); and $e_0$ and the cyclic axial strain amplitude (CSA) (Ottawa F65 sand, all consolidated to about 100\,kPa). The systematic variation of $\ve{\theta}^{(m)}$ enables the representation of diverse mechanical behaviors of sand samples with the same constitutive properties.

\noindent The multiple triaxial tests dataset is represented as follows:
{\small
\begin{align*}
\text{Dataset} = \begin{Bmatrix} \ve{\theta}^{(m)},
    \begin{pmatrix}
        \ve{s}^{(m)}_0 \\
        \ve{u}^{(m)}_0 \\
    \end{pmatrix},
    \ldots,
    \begin{pmatrix}
        \ve{s}^{(m)}_t \\
        \ve{u}^{(m)}_t \\
    \end{pmatrix}, \ldots,
    \begin{pmatrix}
        \ve{s}^{(m)}_{N} \\
        \ve{u}^{(m)}_{N}\\
    \end{pmatrix}
\end{Bmatrix}_{m=1}^M
\end{align*}}
The goal is to characterize the evolution of soil states through the mapping $f_{\beta}$, where a deterministic framework relates the previous state to the next:
\begin{align}
   \ve{s}_{t}^{(m)} = f_{\ve{\beta}} \left(\ve{s}_{t-1}^{(m)}, \ve{u}_t^{(m)}, \ve{\theta}^{(m)}\right), \quad t = 1,\ldots, N, \;\; m=1,\ldots, M
    \label{eq:fnn}
\end{align}
Such a deterministic mapping provides no measure of uncertainty.

To address this, a probabilistic mapping $f_{\beta}$ that relates the previous state to the current state, conditioned on exogenous inputs and triaxial test constants, is proposed. Specifically, given $M$ datasets, each structured as:
\begin{align}
    \mathcal{S}^{(m)} = \cbk{ \ve{u}^{(m)}_{1:{N}}, \ve{s}^{(m)}_{1:{N}}, \ve{\theta}^{(m)} }, \quad m=1,\ldots, M
\end{align}
The objective is to characterize the predictive probability distribution over the states $\ve{s}$, conditioned on the inputs $\ve{u}$ and the test constants $\ve{\theta}$, expressed as:
\begin{align*}
    p \brc{\ve{s}_{1:{N}} \mid \ve{u}_{1:{N}}, \ve{\theta}}
\end{align*}
which encodes both the predictions and the associated uncertainties.

The conditional distribution is modeled as Gaussian, so that only its mean and covariance need to be estimated. The proposed framework uses a generalized variational Bayesian inference integrated with a windowed recursive FFNN structure, to fit the probabilistic representation, quantifying uncertainty across load steps.

\section{Methodology} \label{sec:method}
The proposed framework combines the strengths of deterministic recursive FFNN models with the uncertainty quantification capabilities of Bayesian neural networks. We first introduce rFFNNs and the sliding-window strategy, then extend them into recursive Bayesian FFNNs (rBNNs), which quantify both aleatoric and epistemic uncertainty. Finally, we outline the generalized likelihood, posterior approximation, and normalization used for model training.

\subsection{Recursive Feedforward Neural Networks (rFFNNs)}\label{sec:rFFNN}

Unlike standard FFNNs, which map one state to the next, rFFNNs recursively feed predictions back as inputs, reducing error accumulation over multi-step forecasts because the network is exposed to its own predictions during training (\cref{fig:fbnn}). This recursive loop allows rFFNNs to capture temporal dependencies and is well-suited to sequential constitutive modeling.

\subsubsection{Sliding window approach for training rFFNNs}
\label{sec:sliding}

\begin{figure}[!ht]
     \centering
     \includegraphics[width=0.8\textwidth]{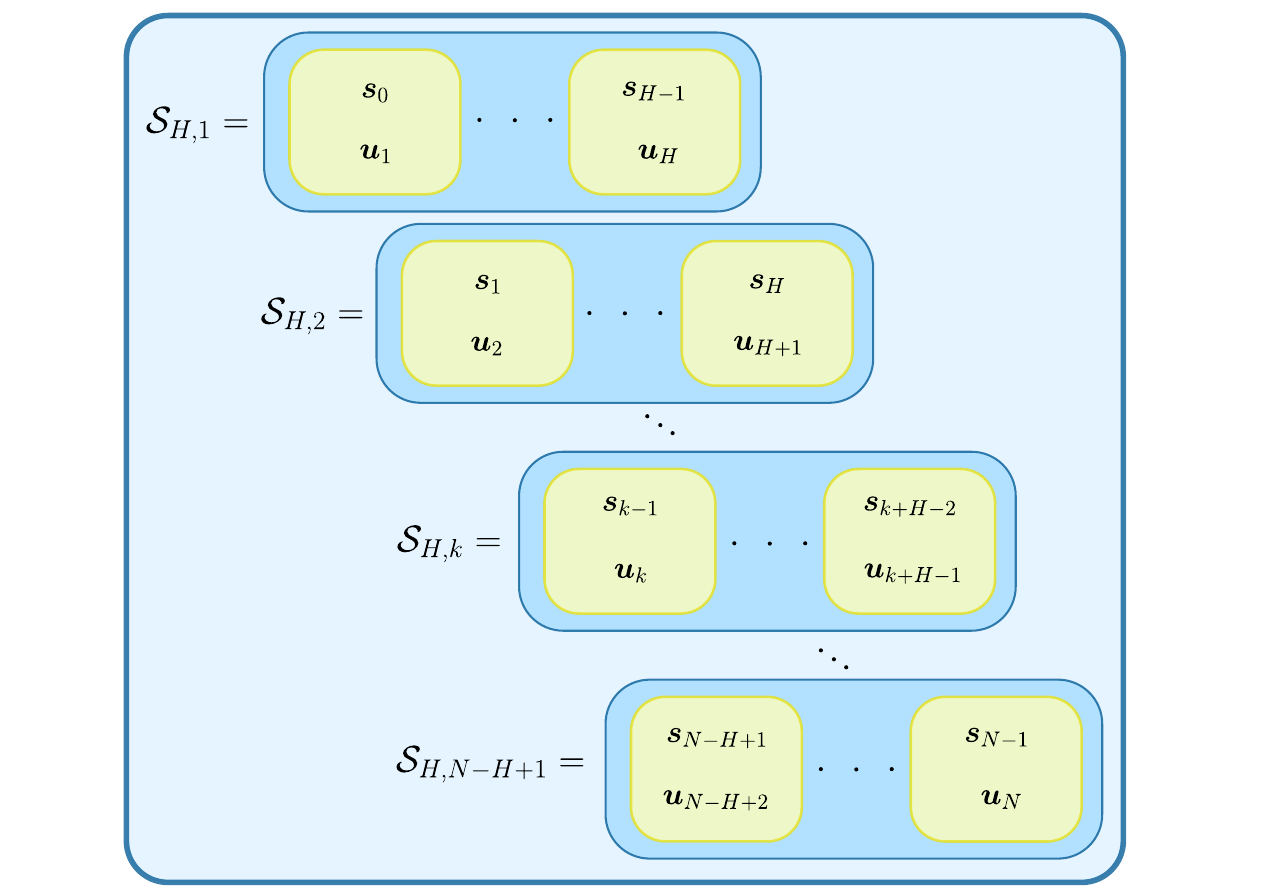}
     \caption{Sliding window segmentation of a triaxial loading sequence into overlapping subsequences of length $H$ for rFFNN training.}
     \label{fig:sample_ffnn_multi}
\end{figure}

The loading series is divided into overlapping subsets or \textit{sliding windows} of length $H$, each containing states, exogenous inputs, and test constants:
\begin{equation}\label{eq:equation_3}
\mathcal{S}_{H,k}^{(m)} = \cbk{\ve{s}_{t-1}^{(m)}, \ve{u}_t^{(m)}, \ve{\theta}^{(m)}}_{t = {k}}^{k + H - 1}, \quad k = 1, \dots, N-H+1, \; m=1,\ldots,M
\end{equation}
Here, $\mathcal{S}_{H,k}^{(m)}$ is a window from the $m^{\text{th}}$ triaxial test, comprising triaxial test constants $\ve{\theta}$ and a certain $H$-long sequence of state variables $\ve{s}$ and exogenous inputs $\ve{u}$. Training on windows (\cref{fig:sample_ffnn_multi}) localizes learning of temporal patterns. $H=1$ reduces to a standard FFNN, while $H=N$ recovers a full-series rFFNN \cite{ellis1995stress}. This flexibility allows the framework to balance the trade-off between capturing short-term patterns and long-term dependencies.

Consecutive windows are shifted by one load step (stride of one), so that each interior load step appears in $H$ windows; this maximizes the number of windows obtained from the limited number of tests, and only $H$ is tuned. One training epoch requires $M(N-H+1)H$ network evaluations, so its cost grows approximately linearly with $H$. Windows are extracted from each test after the split into training, validation and test sets (\cref{sec:assess}), so no window spans two tests or two subsets.

\subsubsection{Recursive prediction within windowed subsets}
\begin{figure}[!ht]
     \centering
     \includegraphics[width=1\textwidth]{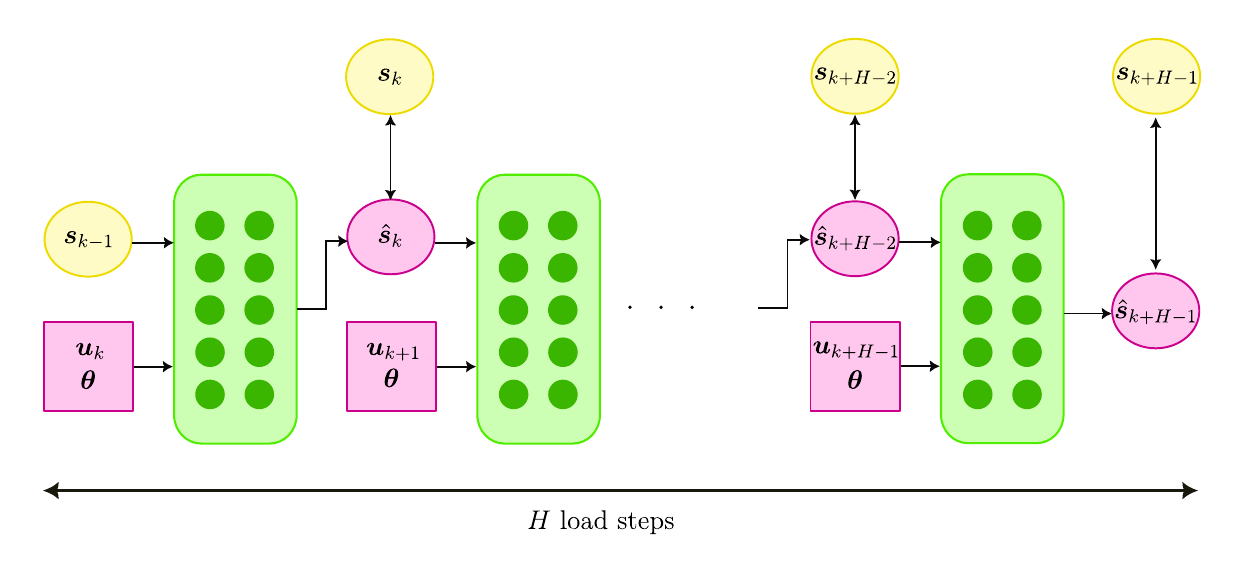}
     \caption{\textbf{Recursive state prediction over a $H$-length sliding window} using a deterministic rFFNN for the data subset \( \mathcal{S}_{H,k} \).}
     \label{fig:sample_0}
\end{figure}

For each window \(\mathcal{S}_{H,k}^{(m)}\), the rFFNN performs recursive predictions starting with the initial measured state \(\ve{s}_{k-1}^{(m)}\) to predict the next state \(\hat{\ve{s}}_{k}^{(m)}\), given the input \(\ve{u}_{k}^{(m)}\) and triaxial test constant \(\ve{\theta}^{(m)}\). The predicted states are recursively fed back into the rFFNN, alongside the inputs from the subsequent load step, to predict the future states within the window. This recursive mechanism, illustrated in \cref{fig:sample_0}, reads:
\begin{align}
\begin{split}
    \hat{\ve{s}}_{k}^{(m)} &= f_{\ve{\beta}}\brc{\ve{s}_{k-1}^{(m)}, \ve{u}_{k}^{(m)}, \ve{\theta}^{(m)}}\\
    \hat{\ve{s}}_{k+1}^{(m)} &= f_{\ve{\beta}}\brc{\hat{\ve{s}}_{k}^{(m)}, \ve{u}_{k+1}^{(m)}, \ve{\theta}^{(m)} }\\
    & \vdots\\
    \hat{\ve{s}}_{k+H-1}^{(m)} &= f_{\ve{\beta}}\brc{\hat{\ve{s}}_{k+H-2}^{(m)}, \ve{u}_{k+H-1}^{(m)}, \ve{\theta}^{(m)}}
\end{split}
\end{align}

\subsubsection{Loss computation for training rFFNNs}
The loss \(L_{H,k}^{(m)}\) for a windowed subset \(\mathcal{S}^{(m)}_{H,k}\) is computed as the average discrepancy between the predicted states \(\hat{\ve{s}}^{(m)}_{t}\) and the corresponding observed states \(\ve{s}_{t}^{(m)}\) across all load steps within the window:
\begin{equation}
L_{H,k}^{(m)} = \frac{1}{H} \sum_{t = k}^{k+H-1} \ell \brc{\ve{s}_{t}^{(m)}, \hat{\ve{s}}_{t}^{(m)}}
\end{equation}
The function $\ell$ is a user-chosen loss metric, such as the squared loss, absolute loss, and Huber loss. In this study the squared loss is used for all deterministic models (\cref{sec:protocol}). Within a window, predicted states are fed back without detaching them, so the gradients propagate through all $H$ recursive steps.

The average loss \(L \) across all loading series of states $\cbk{\ve{s}^{(m)}_{0:N}}_{m=1}^M$, comprising $M$ triaxial tests with $N$ load steps each, is determined by averaging the loss over all windowed subsets, given the complete loading series of control inputs $\cbk{\ve{u}^{(m)}_{1:N}, \ve{\theta}^{(m)}}_{m=1}^M$:

\begin{align}
    L = \sum_{m=1}^M \frac{1}{M(N-H+1)} \sum_{k=1}^{(N-H+1)} L_{H,k}^{(m)}
\end{align}

\subsection{Generalized Bayesian rFFNNs (rBNNs)}\label{sec:rBNN}
\begin{figure}[!ht]
     \centering
     \includegraphics[width=0.8\textwidth]{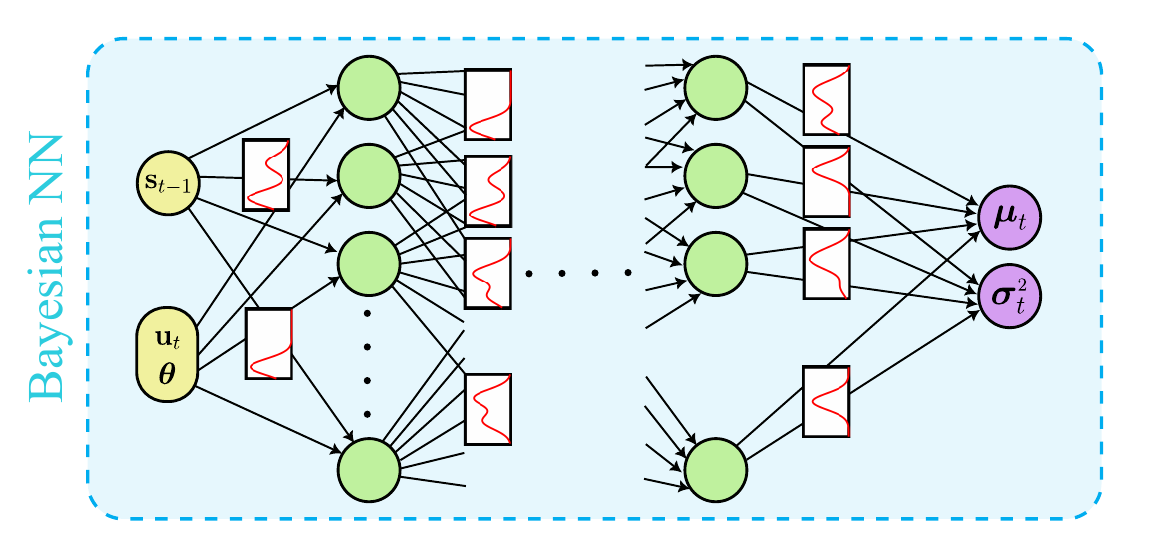}
     \caption{\textbf{Depiction of rBNN} where model parameters are treated as random variables, and the output layer predicts the mean and diagonal variances of a Gaussian distribution for the subsequent state.}
     \label{fig:bnn}
\end{figure}

 The rBNN framework extends rFFNN by treating model parameters as random variables. This approach aligns closely with Bayesian neural networks (BNNs)\cite{neal2012bayesian,gal2016dropout}, which enables the estimation of both epistemic uncertainty (arising from limited training data) and aleatoric uncertainty (due to inherent noise in data). The key enhancement of rBNNs over rFFNNs is as follows:
\begin{enumerate}[leftmargin=2.2em]
    \item[(a)] \textit{Random variable modeling}: The parameter vector $\ve{\beta}$ is a random variable with a joint probability distribution, which captures epistemic uncertainty.

    \item[(b)] \textit{Deterministic output uncertainty layer}: A deterministic layer at the output of the rBNN predicts the mean and covariance of a Gaussian distribution for the next state, capturing aleatoric uncertainty due to noise in the training data.

    \item[(c)] \textit{Use of pseudo-likelihood}: Unlike traditional Bayesian methods, where the likelihood is derived from a probabilistic model of the data, we employ a pseudo-likelihood derived from the recursive predictions over several overlapping subsets of data.
\end{enumerate}
As illustrated in \cref{fig:bnn}, instead of point estimates of $\ve{\beta}$, the posterior distribution over $\ve{\beta}$ is inferred within a \textit{generalized} Bayesian framework.

However, the high dimensionality of parameter vector $\ve{\beta}$ makes posterior inference challenging: analytical solutions are intractable and Markov chain Monte Carlo (MCMC), though exact, is computationally prohibitive. Instead, we adopt variational Bayesian inference, which approximates the posterior with a simpler tractable family. The following subsections present the key components of the generalized Bayesian rFFNN: prior, pseudo-likelihood, posterior approximation, and posterior predictive distribution.

\subsubsection{Prior distribution}

The prior distribution encodes initial beliefs about parameters $\ve{\beta}$ before observing data. A common choice in neural networks is a fully factorized standard Gaussian \cite{fortuin2022priors}, where each $\beta_l \sim \mathcal{N}(0,1)$, due to its simplicity and computational efficiency. Formally,

\begin{equation} \label{eq:prior}
    p(\ve{\beta}) = \prod_{l=1}^{L} p({\beta}_l) = \prod_{l=1}^{L} \mathcal{N}\brc{{\beta}_l; 0, 1} = \mathcal{N} \brc{\ve{\beta}; \ve{0}, \mathbf{I}_L},
\end{equation}
where $L$ denotes the dimensionality of the model parameter vector \( \ve{\beta} \) and \( \mathbf{I}_L \) represents the \( L \times L \) identity matrix. This prior simplifies posterior inference, regularizes the model by favoring parameters near zero, and helps prevent overfitting in high-dimensional settings.

\subsubsection{Pseudo-likelihood formulation} \label{sec:pseudolike}

In classical Bayesian inference, the likelihood is derived from a probabilistic model of the data-generating process. For rFFNNs, however, the sliding-window approach reuses overlapping sequences, complicating a conventional likelihood. To address this, we employ a pseudo-likelihood, which acts like a loss function and efficiently handles overlapping data while enabling uncertainty quantification \cite{matsubara2022robust}.

For the $t^{\text{th}}$ load step in the $k^{\text{th}}$ window of the $m^{\text{th}}$ triaxial test, the conditional distribution of the state vector $\ve{s}_{t}^{(m)}$ is modeled as:
\begin{align} \label{eq:state_conditional}
    p\brc{\ve{s}_{t}^{(m)} \mid \ve{s}_{t-1}^{(m)}, \ve{u}_{t}^{(m)}, \ve{\theta}^{(m)}, \ve{\beta}} = \mathcal{N}\brc{\ve{s}_t^{(m)}; \ve{\mu}_{t}^{(m)}, \ve{\sigma}_t^{2(m)}}
\end{align}
where the predicted mean vector $\ve{\mu}_t^{(m)}$ and the predicted diagonal variance vector $\ve{\sigma}_t^{2(m)}$ are the two outputs of the neural network $f_{\ve{\beta}}$:
{
\begin{align} \label{eq:meancovfun}
    \begin{split}
        \ve{\mu}_t^{(m)}(\ve{\beta}) &= f^{\mu}_{\ve{\beta}} \brc{\ve{s}_{t-1}^{(m)}, \ve{u}_{t}^{(m)}, \ve{\theta}^{(m)}}\\
        \ve{\sigma}_t^{2(m)}(\ve{\beta}) &= f^{\sigma}_{\ve{\beta}} \brc{\ve{s}_{t-1}^{(m)}, \ve{u}_{t}^{(m)}, \ve{\theta}^{(m)}}
    \end{split}
\end{align}}
with $\ve{\beta}$ representing the neural network parameters (weights and biases), and $f^{\mu}_{\ve{\beta}}$ and $f^{\sigma}_{\ve{\beta}}$ the mean and standard-deviation output heads of the same network; positivity of $\ve{\sigma}_t^{(m)}$ is enforced by a softplus function with a floor of $10^{-4}$.

We construct the likelihood for a $H$-length sliding window $\mathcal{S}_{H,k}^{(m)}$ by defining a joint distribution over the state sequence $\ve{s}_{k:k+H-1}^{(m)} = \cbk{\ve{s}_{k}^{(m)}, \ldots, \ve{s}_{k+H-1}^{(m)}}$ for a given triaxial test $m$, with known initial state $\ve{s}_{k-1}^{(m)}$, as a product of conditional distributions:
\begin{align} \label{eq:joint_likelihood_true}
    \mathcal{L}_{H,k}^{(m)}(\ve{\beta}) = p\brc{\ve{s}_{k:k+H-1}^{(m)} \mid \ve{u}_{k:k+H-1}^{(m)}, \ve{\theta}^{(m)}, \ve{\beta}} = \prod_{t=k}^{k+H-1} p \brc{\ve{s}_t^{(m)} \mid \ve{s}_{t-1}^{(m)}, \ve{u}_t^{(m)}, \ve{\theta}^{(m)}, \ve{\beta}}
\end{align}
In the recursive setting, however, only $\ve{s}_{k-1}^{(m)}$ is measured; for $t>k$ the conditioning state is the network's own random prediction, whose full predictive distribution would have to be propagated through the network.
However, propagating the full joint distribution across the load steps is computationally prohibitive, especially for overlapping windows across multiple triaxial tests. To simplify, we approximate the joint likelihood using a pseudo-likelihood formulation, where the state at load $t$ depends on the predicted mean of the previous state rather than its sampled value:
\begin{align} \label{eq:joint_likelihood}
    \begin{split}
        \mathcal{L}_{H,k}^{(m)}(\ve{\beta}) &= p \brc{\ve{s}_k^{(m)} \mid \ve{s}_{k-1}^{(m)}, \ve{u}_k^{(m)}, \ve{\theta}^{(m)}, \ve{\beta}} \prod_{t=k+1}^{k+H-1} p \brc{\ve{s}_t^{(m)} \mid \ve{\mu}_{t-1}^{(m)}, \ve{u}_t^{(m)}, \ve{\theta}^{(m)}, \ve{\beta}} \\
        &= \mathcal{N} \brc{\ve{s}_k^{(m)}; \ve{\mu}_k^{(m)}(\ve{\beta}), \ve{\sigma}_k^{2(m)}(\ve{\beta})} \prod_{t=k+1}^{k+H-1} \mathcal{N} \brc{\ve{s}_t^{(m)} ; \ve{\mu}_t^{(m)}(\ve{\beta}), \ve{\sigma}_t^{2(m)}(\ve{\beta})}
    \end{split}
\end{align}
This approximation reduces computational complexity and memory usage while maintaining key uncertainty quantification features.

Substituting the expression of the Gaussian distribution into the pseudo-likelihood and ignoring terms that are constant with respect to the network parameters $\ve{\beta}$, the log-joint-pseudo-likelihood for the $k^{\text{th}}$ window becomes:
{
\begin{align}\label{eq:equation11}
    \log\mathcal{L}_{H,k}^{(m)}(\ve{\beta}) = - \frac{1}{2H} \sum_{t=k}^{k+H-1} \brc{\log \det \brc{\text{diag}\brc{\ve{\sigma}_t^{2(m)}}} + \brc{ \ve{s}_t^{(m)} - \ve{\mu}_t^{(m)} }^T \cbk{\text{diag}\brc{\ve{\sigma}_t^{2(m)}}}^{-1} \brc{ \ve{s}_t^{(m)} - \ve{\mu}_t^{(m)} } }
\end{align}}
where, as in the rFFNN loss, the terms are averaged over the $H$ steps of the window.
The average log-pseudo-likelihood for the entire sequential loading of states $\mathcal{D} = \cbk{\ve{s}^{(m)}_{0:N}}_{m=1}^M$, consisting of $M$ triaxial tests, each with $N$ load steps, given entire sequential loading series of control inputs $\mathcal{U} = \cbk{\ve{u}^{(m)}_{1:N}, \ve{\theta}^{(m)}}_{m=1}^M$ is computed by averaging the log-pseudo-likelihoods over all windowed subsets:
\begin{align}\label{eq:equation12}
    \log\mathcal{L}(\ve{\beta}) = \log p\brc{\mathcal{D} \mid \ve{\beta}, \mathcal{U}} = \sum_{m=1}^M \frac{1}{M(N-H+1)} \sum_{k=1}^{(N-H+1)} \log\mathcal{L}_{H,k}^{(m)}(\ve{\beta})
\end{align}
where $N-H+1$ is the total number of sliding windows per triaxial test.

\subsubsection{Parameter posterior approximation via generalized variational inference}

In Bayesian inference, the parameter posterior distribution $p\brc{\ve{\beta} \mid \mathcal{D}, \mathcal{U}}$ encapsulates our updated beliefs about the parameters $\ve{\beta}$ after observing the data $\mathcal{D}$. It is intractable because the evidence $p(\mathcal{D} \mid \mathcal{U})$ requires marginalizing over the high-dimensional parameter space; variational inference (VI) approximates it through optimization.

\paragraph{Variational formulation}

Variational inference introduces a family of tractable distributions $q_{\ve{\eta}}(\ve{\beta})$ parameterized by $\ve{\eta}$, which is designed to approximate the true posterior $p(\ve{\beta} \mid \mathcal{D}, \mathcal{U})$. The goal is to find the member of the variational family $q_{\ve{\eta}}(\ve{\beta})$ that minimizes the Kullback-Leibler (KL) divergence -- which measures the information lost in approximation -- to the true posterior:
\begin{align}
D_{\text{KL}} \brc{q_{\ve{\eta}}(\ve{\beta}) \| p(\ve{\beta} \mid \mathcal{D}, \mathcal{U})} = \mathbb{E}_{q_{\ve{\eta}}(\ve{\beta})} \sbk{ \log q_{\ve{\eta}}(\ve{\beta}) - \log p(\ve{\beta} \mid \mathcal{D}, \mathcal{U}) }
\end{align}
Direct minimization of $D_{\text{KL}}$ is not feasible because the true posterior $p(\ve{\beta} \mid \mathcal{D}, \mathcal{U})$ involves the intractable evidence $p(\mathcal{D} \mid \mathcal{U})$. Instead, VI maximizes a surrogate objective called the \textit{Evidence Lower Bound} (ELBO), which is derived from the decomposition of the log evidence, as follows:
\begin{align} \label{eq:logevidence}
\log p(\mathcal{D} \mid \mathcal{U}) = \mathbb{E}_{q_{\ve{\eta}}(\ve{\beta})} \sbk{ \log \frac{p(\mathcal{D}, \ve{\beta} \mid \mathcal{U} )}{q_{\ve{\eta}}(\ve{\beta})} } + D_{\text{KL}} \brc{q_{\ve{\eta}}(\ve{\beta}) \| p(\ve{\beta} \mid \mathcal{D}, \mathcal{U})}
\end{align}
where the second term is the KL divergence, which is always non-negative. This implies:
\begin{align}
\log p(\mathcal{D} \mid \mathcal{U}) \geq \mathbb{E}_{q_{\ve{\eta}}(\ve{\beta})} \sbk{ \log \frac{p(\mathcal{D}, \ve{\beta} \mid \mathcal{U})}{q_{\ve{\eta}}(\ve{\beta})} } \triangleq \mathcal{L}_{\text{ELBO}}
\end{align}
The ELBO can be further decomposed into two terms:
\begin{align} \label{eq:elbo}
\mathcal{L}_{\text{ELBO}} = \mathbb{E}_{q_{\ve{\eta}}(\ve{\beta})} \sbk{\log p(\mathcal{D} \mid \mathcal{U} , \ve{\beta}) } - D_{\text{KL}}\brc{q_{\ve{\eta}}(\ve{\beta}) \| p(\ve{\beta})}
\end{align}
where the first term is the expected log-likelihood and it measures how well the variational distribution $q_{\ve{\eta}}(\ve{\beta})$ fits the observed data, and the second term is the KL divergence from prior that regularizes the variational distribution $q_{\ve{\eta}}(\ve{\beta})$ by penalizing it from deviating too far from the prior $p(\ve{\beta})$. Because the log evidence in \cref{eq:logevidence} does not depend on $\ve{\eta}$, maximizing the ELBO is equivalent to minimizing the KL divergence between $q_{\ve{\eta}}(\ve{\beta})$ and the true posterior.

\paragraph{Monte Carlo approximation of ELBO}

The expectation in the computation of ELBO is approximated using Monte Carlo integration by drawing i.i.d. samples from the variational distribution $q_{\ve{\eta}}(\ve{\beta})$:
\begin{align} \label{eq:approx_elbo}
    \mathbb{E}_{q_{\ve{\eta}}(\ve{\beta})} \sbk{ \log p \brc{\mathcal{D} \mid \ve{\beta}, \mathcal{U}} } \approx \frac{1}{N_q} \sum_{i=1}^{N_q} \log p\brc{\mathcal{D} \mid \ve{\beta}^{(i)}, \mathcal{U}}, \quad \ve{\beta}^{(i)} \sim q_{\ve{\eta}}(\ve{\beta})
\end{align}
where $N_q$ is the number of Monte Carlo samples. A single sample is often sufficient in practice \cite{kucukelbir2017automatic}. In this study, one sample $\ve{\beta}^{(i)}$ is drawn for each training window ($N_q = 1$) and held fixed over its $H$ recursive steps, so that every term of \cref{eq:equation11} is evaluated with the same network, as at prediction (\cref{sec:predictive}). Different windows of a minibatch receive independent samples; because the log-pseudo-likelihood (\cref{eq:equation12}) is an average over windows, this keeps the Monte Carlo estimate unbiased and typically reduces its variance compared with one sample shared by the whole minibatch.

The prior regularization term, $D_{\text{KL}}\brc{q_{\ve{\eta}}(\ve{\beta}) \| p(\ve{\beta})}$, is computed analytically, since the variational distribution $q_{\ve{\eta}}(\ve{\beta})$ and prior $p(\ve{\beta})$ are both Gaussian.

\paragraph{Choice of variational distribution}

 A common, tractable choice for $q_{\ve{\eta}}(\ve{\beta})$ is the fully-factorized Gaussian distribution:
\begin{align} \label{eq:normal_q}
    q_{\ve{\eta}}(\ve{\beta}) = \prod_{i=1}^L \mathcal{N}\brc{\beta_i; \mu_i^q, (\sigma_i^q)^2}
\end{align}
where $\ve{\eta} = \cbk{ \ve{\mu}^q, \ve{\rho}^q}$, with $\ve{\mu}^q$ the mean vector of the variational posterior and $\ve{\rho}^q$ an unconstrained vector that defines its standard deviations through the softplus function, $\ve{\sigma}^q = \log\brc{1+\exp\brc{\ve{\rho}^q}}$. The means $\ve{\mu}^q$ are initialized with the standard (Kaiming-uniform) initialization and all elements of $\ve{\rho}^q$ with a common value $\rho_0 \in \{-6, -5, -4\}$ (initial weight standard deviations of $0.0025$--$0.018$), treated as a hyperparameter.

\paragraph{Optimization of variational parameters}

The variational parameters $\ve{\eta}$ are optimized by maximizing the ELBO using gradient-based optimization. The reparameterization trick \cite{kingma2013auto} is often employed to backpropagate through the sampling process:
{
\begin{align}
    \ve{\beta} = \ve{\mu}^q + \log \brc{1+ \exp \brc{\ve{\rho}^q}} \odot \ve{\gamma}
\end{align}}
where $\ve{\gamma}$ is a noise variable independent of $\ve{\mu}^q$ and $\ve{\rho}^q$, sampled from a standard Gaussian distribution $\mathcal{N}(\ve{\gamma}; \ve{0}, \mathbf{I}_L)$ and $\odot$ represents elementwise multiplication. This parameterization expresses $\ve{\beta}$ as a deterministic function of $\ve{\mu}^q$, $\ve{\rho}^q$ and the random variable $\ve{\gamma}$, allowing flow of gradients via backpropagation through the parameters $\ve{\mu}^q$ and $\ve{\rho}^q$ during optimization. Equivalently, the negative ELBO is minimized with respect to the variational parameters:
{
\begin{align}
\begin{split}
    \ve{\mu}^q_{\star}, \ve{\rho}^q_{\star} &= \arg \min \; - \; \mathcal{L}_{\text{ELBO}} \\
    &= \arg \min \; - \mathbb{E}_{q_{\ve{\eta}}(\ve{\beta})} \sbk{\log p(\mathcal{D} \mid \ve{\beta}, \mathcal{U}) } + \lambda_{\text{KL}}\, D_{\text{KL}}\brc{q_{\ve{\eta}}(\ve{\beta}) \| p\brc{\ve{\beta}}} \\
    & \approx \arg \min \; - \frac{1}{N_q} \sum_{i=1}^{N_q} \log p\brc{\mathcal{D} \mid \ve{\beta}^{(i)}, \mathcal{U}} + \lambda_{\text{KL}}\, D_{\text{KL}}\brc{q_{\ve{\eta}}(\ve{\beta}) \| p\brc{\ve{\beta}}} \\
    & = \arg \min \; - \frac{1}{N_q} \sum_{i=1}^{N_q} \log p\brc{\mathcal{D} \mid \ve{\beta}^{(i)}, \mathcal{U}} + \lambda_{\text{KL}} \sum_{i=1}^L \left[ -\log \sigma_i^q + \frac{1}{2} \left( (\mu_i^q)^2 + (\sigma_i^q)^2 - 1 \right) \right]
\end{split}
\label{eq:neg_elbo}
\end{align}}
where $\ve{\beta}^{(i)} \sim q_{\ve{\eta}}(\ve{\beta})$ and $\lambda_{\text{KL}}$ weights the prior term ($\lambda_{\text{KL}}=1$ recovers the standard ELBO). Because \cref{eq:equation12} averages over windows rather than summing over independent observations, $\lambda_{\text{KL}} = 1/N_w$, with $N_w$ the number of training windows; this is the usual minibatch scaling of the ELBO, in which each window counts as one observation. A smaller weight and R\'enyi divergences \cite{li2016renyi} are examined in the ablation study (\cref{sec:ablation}). In the generalized variational inference view below, $\lambda_{\text{KL}}$ acts as the inverse learning rate (temperature) of the generalized posterior and therefore controls the width of the predictive intervals.
The gradients required to optimize the variational parameters \(\ve{\eta} = [\ve{\mu}^q, \ve{\rho}^q]\) are computed using the Bayes-by-Backprop \cite{blundell2015weight}:

{
\begin{subequations}\label{eq:elbo_grad}
    \begin{align}
    \Delta \ve{\mu}^q &= - \brc{\frac{\partial \mathcal{L}_{\text{ELBO}}}{\partial \ve{\beta}} + \frac{\partial \mathcal{L}_{\text{ELBO}}}{\partial \ve{\mu}^q} } \label{eq:back_1} \\
    \Delta \ve{\rho}^q &= - \brc{ \frac{\partial \mathcal{L}_{\text{ELBO}}}{\partial \ve{\beta}} \odot \frac{\ve{\gamma}}{1 + \exp(-\ve{\rho}^q)} + \frac{\partial \mathcal{L}_{\text{ELBO}}}{\partial \ve{\rho}^q} } \label{eq:back_2}
    \end{align}
\end{subequations}}
where the term \(\frac{\partial \mathcal{L}_{\text{ELBO}}}{\partial \ve{\beta}}\) is computed via standard backpropagation. The overall training procedure is outlined in Algorithm~\ref{alg:training}. Once training is complete, the proposed rBNN model is used recursively to predict over the entire loading sequence, requiring only the state parameters at \( t = 0 \) and exogenous inputs.

\paragraph{Connection to generalized variational inference}

In this work, the expected log-likelihood term in the ELBO (refer \cref{eq:elbo}) is approximated using the pseudo-likelihood formulation described in \cref{sec:pseudolike}. This formulation aligns with generalized variational inference (GVI) as outlined in \cite{knoblauch2019generalized}. GVI extends the classical Bayesian framework to accommodate cases where the likelihood function is replaced by a loss-like function (or a generalized risk function), here the pseudo-likelihood that accommodates the overlapping windows. Consequently, the predictive intervals are not calibrated by construction: each observation enters up to $H$ overlapping windows, and $\lambda_{\text{KL}}$ sets the relative weight of loss and prior. Their calibration is therefore assessed empirically on every test specimen (\cref{sec:calibration}).

\subsection{Posterior prediction} \label{sec:predictive}
Once the optimized variational distribution $q_{\ve{\eta}_{\star}}(\ve{\beta})$ is obtained, the trained rBNN model can predict the complete stress-strain-volume change curve for previously unseen test scenarios. This prediction is conditioned on the initial state parameters $\ve{s}_0^*$ at $t=0$ and the sequence of exogenous inputs $\cbk{\ve{u}^{*}_{1:N_*}, \ve{\theta}^{*}}$, where $N_*$ represents the number of prediction load steps. The predictive probability of the states $\cbk{\ve{s}_{1:N_*}^*}$ is obtained by marginalizing over the posterior weights:
\begin{align}
p(\ve{s}_{1:N_*}^* \mid \ve{u}_{1:N_*}^*, \ve{s}_0^*, \ve{\theta}^*, \mathcal{D}, \mathcal{U})
 = \int p\brc{\ve{s}_{1:N_*}^* \mid \ve{u}_{1:N_*}^*, \ve{s}_0^*, \ve{\theta}^*, \ve{\beta}} \; p \brc{\ve{\beta} \mid \mathcal{D}, \mathcal{U}} \, d\ve{\beta}
\end{align}
To make this computationally feasible, the posterior $p \brc{\ve{\beta} \mid \mathcal{D}, \mathcal{U}}$ is approximated using the optimized variational distribution \( q_{\ve{\eta}_{\star}}(\ve{\beta}) \). The predictive distribution is then approximated via Monte Carlo sampling:
\begin{align}
p \brc{\ve{s}_{1:N_*}^* \mid \ve{u}_{1:N_*}^*, \ve{s}_0^*, \ve{\theta}^*, \mathcal{D}, \mathcal{U}} &= \int p \brc{\ve{s}_{1:N_*}^* \mid \ve{u}_{1:N_*}^*, \ve{s}_0^*, \ve{\theta}^*, \ve{\beta}} \; q_{\ve{\eta}_{\star}}(\ve{\beta}) \, d\ve{\beta} \notag\\
&\approx \frac{1}{N_{\text{mc}}} \sum_{i=1}^{N_{\text{mc}}} p \brc{\ve{s}_{1:N_*}^* \mid \ve{u}_{1:N_*}^*, \ve{s}_0^*, \ve{\theta}^*, \ve{\beta}^{(i)}}, \quad \ve{\beta}^{(i)} \sim q(\ve{\beta}; \ve{\eta}_{\star})
\end{align}
where $N_{\text{mc}} = 100$ denotes the number of Monte Carlo samples. Each sample $\ve{\beta}^{(i)}$ is drawn once and held fixed over the complete loading record, so that it produces one self-consistent trajectory; new weights at every load step would describe only the one-step-ahead uncertainty (ablation variant A9, \cref{sec:ablation}).

Given the Gaussian likelihood structure of the predictive model and the Gaussian variational distribution, the predictive distribution is a uniformly weighted mixture of $N_{\text{mc}}$ Gaussian distributions:
\begin{align}
   p \brc{\ve{s}_{1:N_*}^* \mid \ve{u}_{1:N_*}^*, \ve{s}_0^*, \ve{\theta}^*, \mathcal{D}, \mathcal{U}} &\approx \frac{1}{N_{\text{mc}}} \sum_{i=1}^{N_{\text{mc}}} \mathcal{N} \brc{\ve{s}_{1:N_*}^* \mid \ve{u}_{1:N_*}^*, \ve{s}_0^*, \ve{\theta}^*, \ve{\beta}^{(i)}}, \quad \ve{\beta}^{(i)} \sim \mathcal{N} \brc{\ve{\beta}; \ve{\mu}^q_{\star}, \brc{\ve{\sigma}^q_{\star}}^2} \notag \\
   &= \frac{1}{N_{\text{mc}}} \sum_{i=1}^{N_{\text{mc}}} \prod_{j=1}^{N_*} \mathcal{N}\brc{\ve{s}_j^* \mid \ve{\mu}_j^*\brc{\ve{\beta}^{(i)}}, (\ve{\sigma}_j^*)^2 \brc{\ve{\beta}^{(i)}} }
\end{align}
where $\ve{\mu}^*_j$ and $(\ve{\sigma^*_j})^2$ are the mean and variance of the states at load step $j$, respectively. The recursive nature of the model allows the joint likelihood to be factorized as a product of individual conditional distributions, akin to \cref{eq:joint_likelihood_true}. Consistent with the pseudo-likelihood used in training (\cref{eq:joint_likelihood}), for $j>1$ the state supplied to the network along the $i^{\text{th}}$ sample trajectory is the predicted mean $\ve{\mu}_{j-1}(\ve{\beta}^{(i)})$ of that trajectory.

For each load step, this mixture is summarized by a Gaussian with the same first two moments, which are computed directly from the Monte Carlo samples using the law of total variance:
{
\begin{subequations}\label{eq:pred_moments}
\begin{align}
\ve{\mu}_j^* &= \frac{1}{N_{\text{mc}}} \sum_{i=1}^{N_{\text{mc}}} \ve{\mu}_j \brc{\ve{\mu}_{j-1}(\ve{\beta}^{(i)}), \ve{\theta}^*, \ve{u}_j^*, \ve{\beta}^{(i)}}, \quad \ve{\beta}^{(i)} \sim \mathcal{N} \brc{\ve{\beta}; \ve{\mu}^q_{\star}, \brc{\ve{\sigma}^q_{\star}}^2} \\
(\ve{\sigma}_j^*)^2 &= \underbrace{\frac{1}{N_{\text{mc}}} \sum_{i=1}^{N_{\text{mc}}} \ve{\sigma}_j^2 \brc{\ve{\mu}_{j-1}(\ve{\beta}^{(i)}), \ve{\theta}^*, \ve{u}_j^*, \ve{\beta}^{(i)}}}_{\text{aleatoric}} \; + \; \underbrace{\frac{1}{N_{\text{mc}}} \sum_{i=1}^{N_{\text{mc}}} \brc{\ve{\mu}_j\brc{\cdot\,; \ve{\beta}^{(i)}} - \ve{\mu}_j^*}^2}_{\text{epistemic}} \label{eq:pred_var}
\end{align}
\end{subequations}}
where the squares are taken elementwise, the first term is the expected aleatoric (data-noise) variance, and the second is the epistemic variance arising from the posterior uncertainty in $\ve{\beta}$.
The mean and diagonal variance vectors can then be used to quantify uncertainty. Because the mixture is generally skewed, the 95\% interval is not computed as $\ve{\mu}_j^* \pm 1.96\,\ve{\sigma}_j^*$ but from the 2.5\% and 97.5\% quantiles of the mixture, estimated from 1000 draws per load step (a trajectory $i$ at random, then a value from its Gaussian). The point prediction is the mixture mean $\ve{\mu}_j^*$\footnote{Following common usage in the engineering literature, these intervals of the posterior predictive distribution are referred to as confidence intervals; strictly, they are Bayesian predictive intervals.}.
A small posterior uncertainty in \(\bm{\beta}\) (epistemic) and a small predicted noise variance (aleatoric) lead to narrower confidence intervals, reflecting higher predictive certainty, and vice versa.

\subsection{Data normalization}\label{sec:Data_norm}
Scaling input data is essential in deep learning to avoid ill-conditioning and ensure stable convergence. For soil stress--strain modeling, this is particularly important since stress invariants $p'$ and $q$ are strongly influenced by the initial confining pressure $\sigma'_3$, which governs yield strength and test evolution. 

To achieve this without information from the evaluation specimens, the stresses are first divided by the initial confining stress of each test, $\sigma'_3$: the consolidation pressure of the monotonic tests, the initial mean effective stress $p'_0$ of the simulated cyclic tests, and the measured effective confining stress at the start of cycling, $\sigma'_{3,0}$, of the Ottawa F65 tests. For monotonic loading, where $\sigma'_3$ spans 5--800\,kPa, the stress ratios remain larger at low confinement, so the root-mean-square response of each \textit{training} specimen, $x_{\text{rms}}^{(m)}$, is in addition fitted as a power law of $\sigma'_3$ by least squares in log--log space,
\begin{align}\label{eq:powerlaw}
x_{\text{rms}}^{(m)} \approx a_x \brc{{\sigma'_3}^{(m)}}^{b_x}, \qquad \bar{x}_t^{(m)} = \frac{x_t^{(m)}}{a_x \brc{{\sigma'_3}^{(m)}}^{b_x}}, \qquad x \in \{p', q\},
\end{align}
and the network operates on $\bar{p}'$ and $\bar{q}$. The fitted exponents are slightly below one ($b_p \approx 0.93$, $b_q \approx 0.85$ for both monotonic datasets). Because the scaling factor depends only on $\sigma'_3$, it is available for any new test before it is run.

Other variables---void ratio ($e$), exogenous inputs ($\ve{u}$), and triaxial test constants ($\ve{\theta}$)---and the scaled stresses $\bar{p}'$ and $\bar{q}$ are globally min--max scaled using the extrema over the training tests:
\begin{subequations}\label{eq:min_max}
    \begin{align}
        \bar{x}_t^{(m)} &= \frac{x_t^{(m)} - \underset{t, m}{\min} \; x_t^{(m)} }{\underset{t, m}{\max} \; x_t^{(m)} - \underset{t, m}{\min} \; x_t^{(m)}} \quad \text{for} \quad x \in \{e,u_i\}\\
        \bar{\theta}_i^{(m)} &= \frac{\theta_i^{(m)} - \underset{m}{\min} \; \theta_i^{(m)} }{\underset{m}{\max} \; \theta_i^{(m)} - \underset{m}{\min} \; \theta_i^{(m)}}
    \end{align}
\end{subequations}
The power-law coefficients and the extrema are computed from the training tests only (for the final models, from the training and validation tests, \cref{sec:protocol}); no measured response of a test specimen is used at any stage.

\begin{algorithm}[!ht]
\caption{Training procedure for rBNN}
\label{alg:training}
\begin{algorithmic}[1]
\Require Training dataset \(\mathcal{D} = \cbk{\ve{s}^{(m)}_{0:N}}_{m=1}^M, \mathcal{U} = \cbk{\ve{u}^{(m)}_{1:N}, \ve{\theta}^{(m)}}_{m=1}^M \), Sliding window length $H$
\State Data preprocessing
\begin{itemize}[leftmargin=*]
    \item Normalize the stress invariants $p'$ and $q$ using $\sigma'_3$ and, for monotonic loading, the power law of \cref{eq:powerlaw} fitted on the training tests, and all variables using min-max scaling (\cref{eq:min_max})
    \item Using the normalized input data, create \(M (N - H + 1)\) windowed subsets of data \(\mathcal{S}_{H,k}^{(m)}\) according to \cref{eq:equation_3}
\end{itemize}
\State Define model parameter prior: $\ve{\beta} \sim \mathcal{N}(\ve{\beta}; \ve{0}, \mathbf{I}_L)$
\State Initialize variational parameters $\ve{\mu}^q$ and $\ve{\rho}^q$ (using random initialization)
\For{each minibatch of windows, until the validation-based stopping criterion of \cref{sec:protocol} is met:}
    {
    \For{each window $\mathcal{S}_{H,k}^{(m)}$ in the minibatch (in parallel):}
        \State Sample $\ve{\gamma} \sim \mathcal{N}(\ve{\gamma}; \ve{0}, \mathbf{I}_L)$ and set $\ve{\beta} = \ve{\mu}^q + \log(1+\exp(\ve{\rho}^q)) \odot \ve{\gamma}$ (one sample per window)
        \State Set the input state to the measured $\ve{s}_{k-1}^{(m)}$
        \For{$t = k$ to $k+H-1$:}
            \State Compute $\ve{\mu}_t^{(m)}$ and $\ve{\sigma}_t^{(m)}$ with this $\ve{\beta}$ (\cref{eq:meancovfun}), add the step's term of \cref{eq:equation11} and feed back $\ve{\mu}_t^{(m)}$
        \EndFor
    \EndFor}
    \State Approximate the ELBO from \cref{eq:elbo} as the average log-pseudo-likelihood of the minibatch windows (\cref{eq:equation12}) minus $\lambda_{\text{KL}}$ times the closed-form KL divergence
    \State Compute the gradients of the negative ELBO w.r.t.\ variational parameters \(\Delta \ve{\mu}^q\) and \(\Delta \ve{\rho}^q\) using \cref{eq:elbo_grad}
    \State Update the variational parameters using Adam optimizer with learning rate \(\lambda\) (shown below as a plain gradient step for clarity):
    {
    \begin{align*}
        \ve{\mu}^q &\gets \ve{\mu}^q - \lambda \cdot \Delta \ve{\mu}^q \\
        \ve{\rho}^q &\gets \ve{\rho}^q - \lambda \cdot \Delta \ve{\rho}^q
    \end{align*}}
\EndFor
\State \textbf{Output:} Converged values of variational parameters $\ve{\eta}_{\star} = \cbk{ \ve{\mu}^q_{\star}, \ve{\rho}^q_{\star}}$
\end{algorithmic}
\end{algorithm}

\section{Performance assessment} \label{sec:assess}

This section evaluates deep learning models on monotonic and cyclic triaxial test data. For each loading condition, two datasets are used: numerically simulated tests for benchmarking and experimental tests on sand specimens for real-world validation. 

\begin{figure}[!ht]
    \centering
    \includegraphics[width=0.92\textwidth]{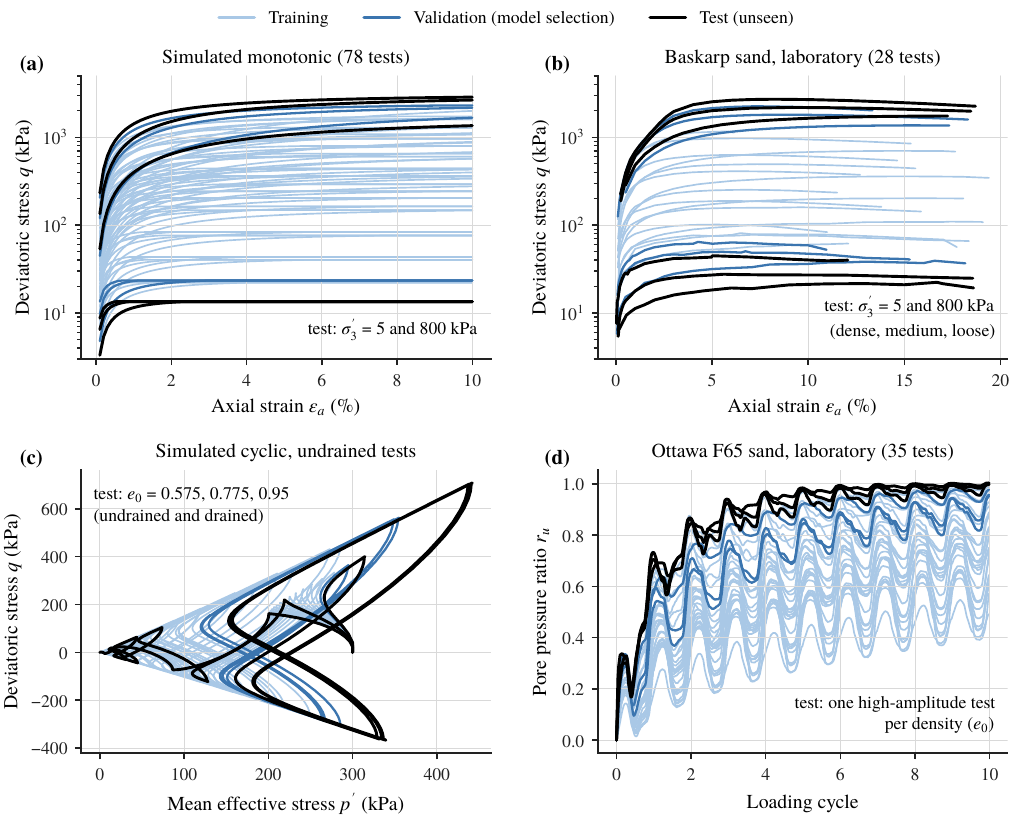}
    \caption{\textbf{The four datasets and their test-level splits}: (\textbf{a}) simulated monotonic drained tests and (\textbf{b}) drained tests on Baskarp sand, deviatoric stress versus axial strain (logarithmic stress axis); (\textbf{c}) simulated cyclic tests, effective stress paths of the undrained tests; (\textbf{d}) undrained strain-controlled cyclic tests on Ottawa F65 sand, excess pore pressure ratio versus loading cycle. Light blue: training tests; dark blue: validation tests; black: test specimens, which are not used for training or model selection.}
    \label{fig:datasets}
\end{figure}

The monotonic datasets include 78 simulated (in \cref{sec:Simulation_Data_mono}) and 28 experimental CD tests (\cref{sec:Experimental_Data_mono}), while the cyclic datasets comprise 32 simulated (drained and undrained) (in \cref{sec:Simulation_Data_cycle}) and 35 experimental CU tests (in \cref{sec:Experimental_Data_cycle}). Each dataset was split into training, validation, and testing sets, with validation guiding hyperparameter selection. All splits are made at the level of whole tests, and the sliding windows are extracted only after this assignment, so no part of a validation or test specimen is seen in training. \Cref{tab:splits} summarizes the splits and \cref{fig:datasets} shows the datasets. The evaluation specimens probe generalization beyond the training conditions: they lie at the extreme confining pressures (monotonic datasets), at initial void ratios outside or between the training values (simulated cyclic), and at the largest strain amplitudes of two of the three density groups (Ottawa F65). In the experimental datasets, they share initial void ratios with training specimens but are physically distinct specimens.

{
\begin{table}[!ht]

\footnotesize
\caption{\textbf{Test-level data splits} for the four datasets. ``Extrap.'' and ``Interp.'' indicate whether the evaluation conditions lie outside or inside the range spanned by the training tests. Validation tests serve only for model selection; the final models are retrained on the training and validation tests. Test specimens are never used for training or selection.}
\label{tab:splits}
\begin{tabularx}{\textwidth}{>{\raggedright\arraybackslash\hsize=0.85\hsize}X>{\raggedright\arraybackslash\hsize=0.95\hsize}X>{\raggedright\arraybackslash\hsize=1.15\hsize}X>{\raggedright\arraybackslash\hsize=0.95\hsize}X>{\raggedright\arraybackslash\hsize=1.10\hsize}X}
\toprule
\textbf{Dataset} & \textbf{Varied conditions} & \textbf{Training} & \textbf{Validation} & \textbf{Test} \\
\midrule
Simulated monotonic (78 CD tests) & $\sigma'_3$ (13 levels, 5--800\,kPa); $E_{50}^{\text{ref}} = 10, 28, 50$\,MPa; $\phi = 30^\circ, 40^\circ$ & 66 tests: all other specimens, including the remaining parameter combinations at 5, 10, 640 and 800\,kPa & 6 tests: $\sigma'_3 = 10$\,kPa ($\phi = 30^\circ$) and 640\,kPa ($\phi = 40^\circ$) & 6 tests: $\sigma'_3 = 5$\,kPa ($\phi = 30^\circ$) and three of the six tests at 800\,kPa (new combinations of $\sigma'_3$, $E_{50}^{\text{ref}}$, $\phi$) \\
\addlinespace[2pt]
Simulated cyclic (16 drained + 16 undrained) & $e_0$; drainage condition & 24 tests: 12 values of $e_0 \in [0.625, 0.925]$ per drainage condition & 2 tests: $e_0 = 0.600$ (Extrap.) & 6 tests: $e_0 = 0.575$ (Extrap.), $0.775$ (Interp.), $0.950$ (Extrap.), both drainage conditions \\
\addlinespace[2pt]
Experimental monotonic, Baskarp (28 CD tests) & $\sigma'_3$ (10 levels, 5--800\,kPa); $e_0 \approx 0.61, 0.70, 0.85$ & 16 tests: $\sigma'_3 = 20$--$320$\,kPa, all three densities & 6 tests: $\sigma'_3 = 10$ and 640\,kPa & 6 tests: $\sigma'_3 = 5$ and 800\,kPa, all three densities (Extrap. in $\sigma'_3$) \\
\addlinespace[2pt]
Experimental cyclic, Ottawa F65 (35 CU tests) & CSA (0.04--0.30\%); $e_0 = 0.576, 0.608, 0.668$ & 29 tests: all other specimens & 3 tests: CSA $= 0.15$, $0.125$, $0.10\%$ (Interp.) & 3 tests: CSA $= 0.25\%$ ($e_0 = 0.576$, Interp.), $0.22\%$ ($0.608$, Extrap.), $0.14\%$ ($0.668$, Extrap.) \\
\bottomrule
\end{tabularx}
\end{table}}

\subsection{\texorpdfstring{Benchmark models and training protocol}{Benchmark models and training protocol}}\label{sec:protocol}
\textit{Models.} Eight models are compared. The FFNN, rFFNN and rBNN share one step architecture, a multilayer perceptron mapping $(\ve{s}_{t-1}, \ve{u}_t, \ve{\theta})$ to $\ve{s}_t$; the FFNN is trained on one-step transitions ($H = 1$) but, like all step models, applied recursively at test time. Two step models with a Gaussian output serve as uncertainty baselines: the \textit{deep ensemble} \cite{lakshminarayanan2017simple}, rFFNNs with a mean and a standard-deviation output trained with the Gaussian negative log-likelihood (NLL) on the same windows as the rBNN, which therefore differ from the rBNN only in their point-estimate weights; and \textit{MC dropout} \cite{gal2016dropout}, the same network with dropout after every hidden layer and one dropout mask per predicted trajectory, analogous to one weight sample per rBNN trajectory. The sequence baselines are the LSTM, GRU and encoder--decoder (\cref{sec:intro}, \cref{appendix:implementation}).

\textit{Information available to the models.} All models share the inputs, splits, normalization and inference protocol (\cref{tab:comparability}): each test curve is predicted from the measured initial state $\ve{s}_0$, the test constants $\ve{\theta}$ and the prescribed strain inputs $\ve{u}_{1:N}$ only, without any measured state after $t=0$. The recurrent models receive $\ve{s}_0$ concatenated with $\ve{u}_t$ at every step. For the simulated monotonic tests, drained loading at constant $\sigma'_3$ implies $p' = \sigma'_3 + q/3$, so $p'$ is derived from the predicted $q$ for all models. All metrics are computed by one script from the saved predictions.

\textit{Training.} All models were implemented in PyTorch \cite{paszke2019pytorch} and trained with Adam \cite{kingma2015adam}. The learning rate was halved after four validation checks without improvement, the gradient norm was clipped at one, and training stopped after $10\,000$ updates or after ten checks (one every 200 updates) without improvement of the validation RMSE of the full recursive prediction, restoring the best parameters. Losses and training sequences are listed in \cref{tab:comparability}; the recurrent models are trained on random segments as long as the shortest training record (the complete records, except for the Ottawa F65 tests, whose records differ in length), starting from the measured state at the segment start.

\textit{Equal hyperparameter search.} For every model and dataset, 12 configurations were drawn at random \cite{bergstra2012random} from search spaces of equal size. Step models: hidden width $\{64, 128, 256\}$, 2 or 3 hidden layers, learning rate $\{3\times10^{-4}, 10^{-3}, 3\times10^{-3}\}$, ReLU or SiLU activation, with or without residual connections, weight decay $\{0, 10^{-5}\}$, batch size $\{128, 256, 512\}$ and window length $H \in \{2, 4, 6, 9, 12, 16\}$ (simulated monotonic), $\{3, 6, 9, 12, 14, 16\}$ (simulated cyclic), $\{2, 4, 6, 8, 11, 14\}$ (Baskarp) or $\{3, 5, 7, 9, 11, 13, 16\}$ (Ottawa F65); in addition, the dropout rate $\{0.05, 0.1, 0.2\}$ for MC dropout and $\rho_0 \in \{-6, -5, -4\}$ for the rBNN. Recurrent models: hidden width $\{32, 64, 128\}$, 1 or 2 layers, the same learning rates and weight decays, dropout $\{0, 0.1\}$ and batch size $\{4, 8, 16\}$. The configurations were ranked on the validation tests by the continuous ranked probability score (CRPS, \cref{sec:metrics}) for the probabilistic models and by the RMSE for the point models, both on the normalized scale. \Cref{tab:hyper} lists the best configurations; Supplementary Fig.~S7 shows the validation scores against $H$.

\textit{Final models.} The five best configurations of every model were retrained on the training and validation tests for their best number of updates, and those with a validation score within a factor of two of the best were retained (one to five per model; Supplementary Table~S2). Their predictions are pooled with equal weights, as the mean (point models) or as one mixture (probabilistic models). This hyperparameter ensemble \cite{wenzel2020hyperparameter} is applied identically to all eight models; the pooled variance-output rFFNNs form the deep ensemble. Each configuration was trained once with a fixed random seed.

{
\begin{table}[!ht]

\footnotesize
\caption{\textbf{Comparability of the benchmarked models.}}
\label{tab:comparability}
\begin{tabularx}{\textwidth}{>{\raggedright\arraybackslash\hsize=0.78\hsize}X>{\raggedright\arraybackslash\hsize=1.12\hsize}X>{\raggedright\arraybackslash\hsize=1.0\hsize}X>{\raggedright\arraybackslash\hsize=1.12\hsize}X>{\raggedright\arraybackslash\hsize=0.98\hsize}X}
\toprule
 & \textbf{LSTM / GRU / Enc.--dec.} & \textbf{FFNN / rFFNN} & \textbf{Deep ensemble / MC dropout} & \textbf{rBNN} \\
\midrule
Inputs at step $t$ & $\ve{u}_t$, $\ve{\theta}$, $\ve{s}_0$ & $\hat{\ve{s}}_{t-1}$, $\ve{u}_t$, $\ve{\theta}$ & $\ve{\mu}_{t-1}$, $\ve{u}_t$, $\ve{\theta}$ & $\ve{\mu}_{t-1}(\ve{\beta}^{(i)})$, $\ve{u}_t$, $\ve{\theta}$ \\
Information at inference & \multicolumn{4}{l}{Identical for all models: $\ve{s}_0$, $\ve{\theta}$ and $\ve{u}_{1:N}$; no measured state after $t=0$} \\
Splits, normalization & \multicolumn{4}{l}{Identical for all models (\cref{tab:splits}, \cref{sec:Data_norm})} \\
Training sequences & Random segments (length of the shortest training record) & Windows of length $H$ ($H = 1$ for FFNN) & Windows of length $H$ & Windows of length $H$ \\
Training loss & Squared error & Squared error & Gaussian NLL & Negative ELBO \\
Hyperparameter search & \multicolumn{4}{l}{12 random configurations from a search space of equal size} \\
Model selection & Validation RMSE & Validation RMSE & Validation CRPS & Validation CRPS \\
Final prediction & \multicolumn{4}{l}{Pool of the retained configurations, retrained on training and validation tests} \\
Uncertainty & -- & -- & Mixture over members; 100 dropout masks per member & 100 weight samples per member \\
Framework & \multicolumn{4}{l}{PyTorch, one code base and one metric script for all models} \\
\bottomrule
\end{tabularx}
\end{table}}

\input{tables/hyper}

\subsection{\texorpdfstring{Evaluation metrics}{Evaluation metrics}}\label{sec:metrics}
Model performance on the validation and testing set (denoted by subscript $*$) was evaluated using three key error metrics: RMSE, MAE and the coefficient of determination $R^2$. These metrics were computed on the predicted states, which could either be scalar components ($p_*$, $q_*$, or $e_*$) or full vector-valued $\ve{s}_* = \sbk{p_*, q_*, e_*}$. With $\hat{\ve{s}}_{t*}^{(m)}$ and $\ve{s}_{t*}^{(m)}$ denoting the predicted and observed values, respectively, of the state variable at load-step $t$ for the $m^{\text{th}}$ experimental sample, the metrics are defined as follows:
{
\begin{subequations}\label{eq:test_metric}
    \begin{align}
        \text{RMSE} &= \sqrt{\frac{1}{M_* N_*} \sum_{m=1}^{M_*} \sum_{t=1}^{N_*} \norm{\ve{s}_{t*}^{(m)} - \hat{\ve{s}}_{t*}^{(m)}}_2^2}, \qquad
        \text{MAE} = \frac{1}{M_* N_*} \sum_{m=1}^{M_*} \sum_{t=1}^{N_*} \norm{\ve{s}_{t*}^{(m)} - \hat{\ve{s}}_{t*}^{(m)}}_1, \\
        R^2 &= 1 - \frac{\sum_{m}\sum_{t} \brc{s_{t*}^{(m)} - \hat{s}_{t*}^{(m)}}^2}{\sum_{m}\sum_{t} \brc{s_{t*}^{(m)} - \bar{s}_*}^2}
    \end{align}
\end{subequations}}
Here, $M_*$ denotes the total number of triaxial tests in the evaluation (validation or test) set, and $N_*$ denotes the number of load steps in the evaluation sets, and $\bar{s}_*$ is the mean of the observed values. In the case of scalar-valued predictions (e.g. $p_*$, $q_*$, $e_*$) the vector norms reduce to scalar operations. RMSE and MAE of the stresses are reported in kPa. $R^2$ is computed for each state variable separately on the $\sigma'_3$-normalized stresses ($p'/\sigma'_3$, $q/\sigma'_3$), so that the large range of confining pressures does not inflate it; the mean $R^2$ averages the three state variables.

The predictive intervals of the three probabilistic models are assessed for each state variable from the 2.5\% and 97.5\% quantiles $l_{t*}^{(m)}$ and $u_{t*}^{(m)}$ of the predictive mixture (\cref{sec:predictive}) by the prediction interval coverage probability (PICP), the fraction of observed values inside the interval; the normalized interval width (PINAW), the mean width divided by the range $R$ of the observed test values; and the interval score \cite{gneiting2007strictly}, which adds a penalty for every observation outside the interval to the width,
\begin{align}\label{eq:calib_metric}
\text{IS}_{0.05} = \frac{1}{R\,M_* N_*}\sum_{m=1}^{M_*}\sum_{t=1}^{N_*} \sbk{\brc{u_{t*}^{(m)} - l_{t*}^{(m)}} + \frac{2}{0.05}\brc{l_{t*}^{(m)} - s_{t*}^{(m)}}_{+} + \frac{2}{0.05}\brc{s_{t*}^{(m)} - u_{t*}^{(m)}}_{+}},
\end{align}
where $(\cdot)_+ = \max(\cdot, 0)$. Calibrated intervals have $\text{PICP} = 0.95$; lower values indicate over-confident and higher values conservative intervals. The interval score rewards narrow intervals only if they contain the observations. Because a pooled coverage can hide a poorly covered specimen, a test specimen is called \textit{reliable} if its 95\% interval contains at least 90\% of its measured points for each of its three state variables. Reliability diagrams (achieved against nominal coverage for nominal levels of 0.05--0.95) are given in Supplementary Fig.~S5. The CRPS used for model selection is computed from 1000 samples of the predictive mixture \cite{gneiting2007strictly}.

Because inference is always performed autoregressively over the complete test sequence (up to 1040 steps for the Ottawa F65 tests), irrespective of the training window length, the test errors directly measure long-horizon stability; the contrast between the FFNN ($H=1$) and the rFFNN isolates the effect of training over recursive windows.

\Cref{tab:accuracy} lists the test accuracy of all models on all four datasets, and \cref{tab:uncertainty} the quality of the predictive intervals of the three probabilistic models; the datasets are discussed in turn below.

\input{tables/accuracy}

\subsection{Evaluation on simulated data}\label{sec:Simulation_Data}
\subsubsection{Monotonic Loading}\label{sec:Simulation_Data_mono}
Synthetic stress--strain--volume change data were generated through CD triaxial tests simulated in PLAXIS 3D using the hardening soil model \cite{schanz1999hardening}. Model parameters were chosen to reflect realistic sand behavior: the internal friction angle $\phi = 30^\circ$ for loose and $40^\circ$ for dense sand, and the reference secant stiffness modulus \( E_{50}^{\text{ref}} \) = 10, 28 and 50 MPa \cite{wichtmann2017correlations}. A constant dilatancy angle \( \psi \) of $5\degree$ was applied. A cohesion of 1 kPa ensured numerical stability. Representative stress--strain responses under varying confining pressures are shown in \cref{fig:datasets}a.

The dataset comprised 78 tests with $N=100$ strain increments (10\% axial strain) across confining pressures 5, 10, 20, 40, 80, 100, 120, 180, 220, 300, 480, 640, and 800 kPa; the six tests at each confining pressure correspond to the six combinations of $\phi$ and $E_{50}^{\text{ref}}$. Data were split into training, validation and test sets as listed in \cref{tab:splits}: the test set contains the three tests with $\phi = 30^\circ$ at 5\,kPa and three of the six tests at 800\,kPa, and the validation set the corresponding tests at 10 and 640\,kPa. Because tests with the other parameter combinations at 5 and 800\,kPa are in the training set, this split assesses generalization to new combinations of $\sigma'_3$, $E_{50}^{\text{ref}}$ and $\phi$ at the extreme pressures rather than extrapolation in $\sigma'_3$ alone. In a preliminary run with a stricter split, in which all twelve tests at 5 and 800\,kPa were held out, all models lost accuracy at the extreme pressures and the 95\% intervals of all three probabilistic models covered only about half of the test points; extrapolating the stress level beyond the training range remains difficult for every model considered here.
The best configurations used $H = 4$ for the rFFNN and $H = 9$ for the rBNN (\cref{tab:hyper}).

All models were trained with the common protocol of \cref{sec:protocol}.

\begin{figure}[!ht]
    \centering
    \includegraphics[width=0.92\textwidth]{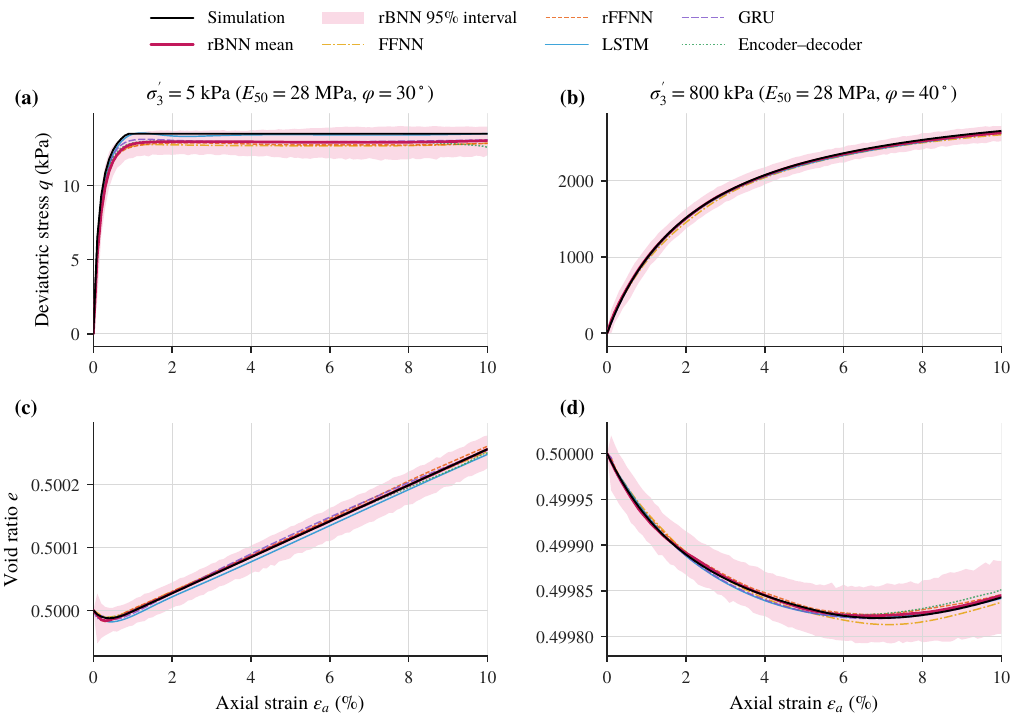}
        \caption{Comparison of model predictions for \textbf{deviatoric stress} $q$ and \textbf{void ratio} $e$ versus \textbf{axial strain} \ensuremath{\varepsilon} at two extreme confining pressures in simulated CD tests: (\textbf{a}, \textbf{c}) $\sigma'_3 = 5$\,kPa, $E_{50}^{\text{ref}} = 28$\,MPa, $\phi = 30^\circ$; (\textbf{b}, \textbf{d}) $\sigma'_3 = 800$\,kPa, $E_{50}^{\text{ref}} = 28$\,MPa, $\phi = 40^\circ$. Each subplot overlays predictions from various ML models, with the rBNN output including 95\si{\percent} confidence intervals; $p' = \sigma'_3 + q/3$ is not shown.}
    \label{fig:1_graphs}
\end{figure}

All models reproduce the simulated responses closely (mean $R^2 \ge 0.985$, \cref{tab:accuracy}). The LSTM has the highest $R^2$ (0.999) and the GRU the lowest stress errors (RMSE of 2.4 and 7.1\,kPa for $p'$ and $q$); the rBNN (2.7 and 8.0\,kPa, mean $R^2 = 0.992$) is close to the deep ensemble and the encoder--decoder. Because the RMSE in kPa is dominated by the 800\,kPa tests, the $R^2$ of the normalized stresses better reflects the low-pressure tests, where the rBNN slightly underestimates the deviatoric stress plateau at 5\,kPa (by about 4\%, \cref{fig:1_graphs}a); the LSTM reproduces it more closely. Predictions under extreme pressures (5 and 800 kPa) are illustrated in \cref{fig:1_graphs}; all six test specimens are shown in Supplementary Fig.~S1.
In these simulations the void ratio changes by less than $10^{-3}$ (\cref{fig:1_graphs}c,d); all models reproduce the dilative trend at 5\,kPa and the contraction followed by dilation at 800\,kPa.

Although LSTM gave the most accurate point predictions, rBNN provided uncertainty-aware forecasts whose 95\% intervals contain 98\% of the simulated stresses and all void ratios, with all six test specimens reliable (\cref{tab:uncertainty}); the intervals are narrow at 800\,kPa and wider at 5\,kPa, where the simulated plateau lies close to their upper edge.

\subsubsection{Cyclic Loading}\label{sec:Simulation_Data_cycle}

To simulate the cyclic response of sandy soils, a synthetic dataset was generated using the exponential constitutive model \cite{yin2018modeling}, which captures stiffness degradation, strain accumulation, and phase transformation under drained and undrained triaxial conditions (see \cref{eq:yineq1}):
\begin{equation}
\dot{\varepsilon}_v = M_{pt} \left| \dot{\varepsilon}_d \right| - \frac{q}{p'} \dot{\varepsilon}_d, \qquad
\dot{q} = 3G \left[ \dot{\varepsilon}_d - \left( \frac{q}{p' M_p} \right) \left| \dot{\varepsilon}_d \right| \right]
\label{eq:yineq1}
\end{equation}
Here \( \dot{q} \) is the deviatoric stress rate, \( \dot{\varepsilon}_v \) the volumetric strain rate, and \( \dot{\varepsilon}_d \) the deviatoric strain rate. In drained conditions, volumetric strain evolves freely. Under undrained loading, \( \dot{\varepsilon}_v = 0 \), producing pore pressure changes. Shear modulus \( G \) is nonlinear and stress-dependent, degrading with strain and void ratio. Strength parameters \( M_p \) and \( M_{pt} \) evolve with state variables, consistent with the critical state concept, and a nonlinear stress--dilatancy law simulates transitions between contractive and dilative responses. Model parameters follow \cite{yin2018modeling} (Table~\ref{tab:yin_params}).

\begin{table}[!ht]
\footnotesize
\caption{Constitutive model parameters adapted from \protect\cite{yin2018modeling}}
\label{tab:yin_params}
\begin{tabularx}{\textwidth}{CCCCCCCC}
\toprule
\( G_0 \) (kPa) & \( \nu \) & \( n \) & \( \phi_c \) (\degree) & \( e_{c0} \) & \( \lambda \) & \( \xi \) & \( d \) \\
\midrule
3000 & 0.3 & 0.67 & 31.2 & 0.937 & 0.022 & 0.71 & 2 \\
\bottomrule
\end{tabularx}
\end{table}

Simulations included drained and undrained conditions, each with 16 strain-controlled cyclic triaxial tests. The initial void ratio \( e_0 \) was varied: 12 values for training (ranging from \(0.625\) to \(0.925\)), one for validation (\(0.600\)), and three for testing (\(0.575,0.775,0.950\)). Undrained tests used confining pressure of $300$ kPa with axial strain amplitude (\(-1\%\)--\(1\%\)); drained tests used $98$ kPa with axial strain amplitude (\(-1.33\%\)--\(1.33\%\)). Each test produced a full cyclic stress--strain path of 300 load steps (three cycles). For the undrained tests, the total volumetric strain is zero, and the third state variable $\varepsilon_v$ is the accumulated inelastic (plastic) volumetric strain given by the stress--dilatancy law; it governs the decrease of the mean effective stress, $\dot{p}' = -K \dot{\varepsilon}_v^{\text{in}}$, with $K$ the bulk modulus.

The best configurations used $H = 16$ for the rFFNN and $H = 14$ for the rBNN (\cref{tab:hyper}), longer windows than for monotonic loading.

\begin{figure}[!ht]
    \centering
    \includegraphics[width=0.92\textwidth]{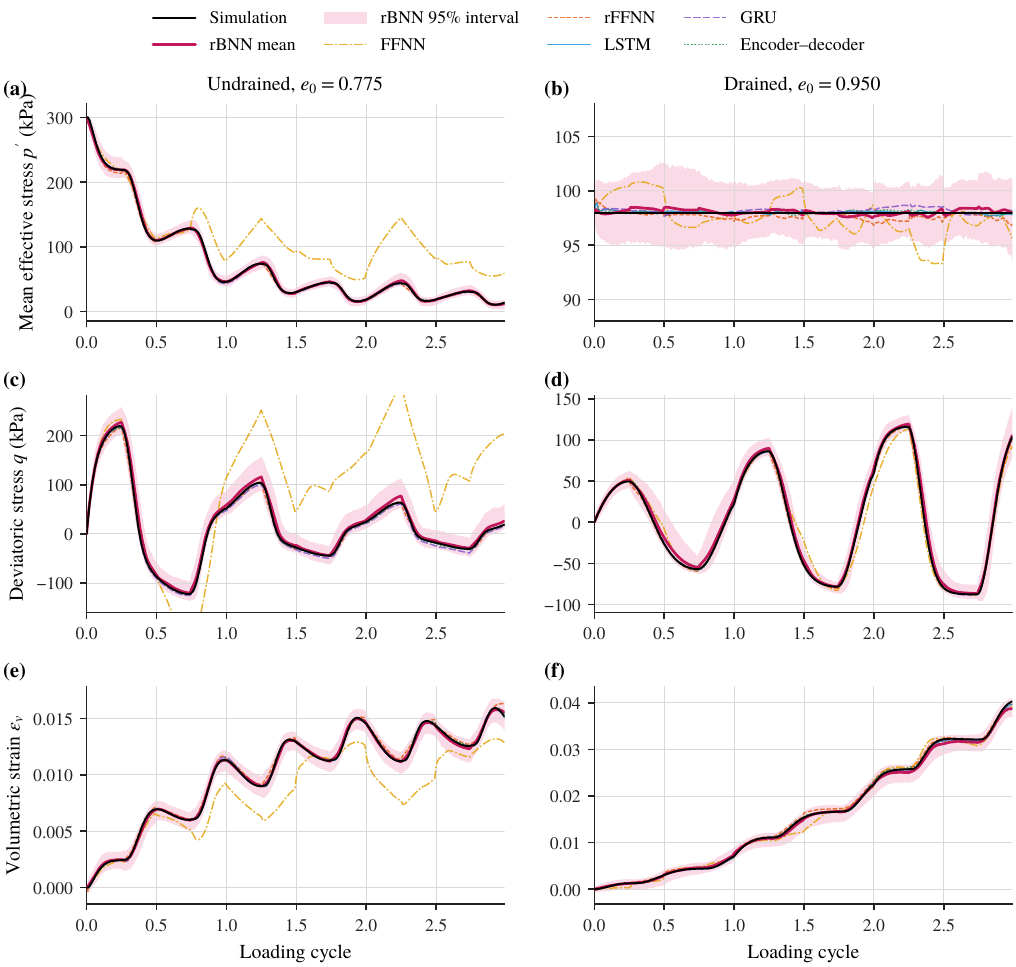}
    \caption{Comparison of model predictions on the test set for \textbf{mean effective stress} $p'$, \textbf{deviatoric stress} $q$ and \textbf{volumetric strain} $\varepsilon_v$ (inelastic volumetric strain in the undrained test) versus \textbf{loading cycle} in simulated cyclic triaxial tests. Left column corresponds to an \textbf{undrained test} with initial void ratio $e_0 = 0.775$, and right column to a \textbf{drained test} with $e_0 = 0.950$. Each subplot overlays predictions from various ML models, with the rBNN output including 95\% confidence intervals.}
    \label{fig:2_graphs_cycle}
\end{figure}

Under cyclic loading, one-step training fails: the FFNN drifts away from the simulated response after the first unloading of the undrained tests (\cref{fig:2_graphs_cycle}a,c; mean $R^2 = 0.892$), whereas all models trained on recursive windows or complete sequences remain accurate over the three cycles (mean $R^2$ of 0.995--0.998; MC dropout 0.979). The LSTM and GRU are the most accurate; the rBNN (0.995) is on par with the rFFNN, the deep ensemble and the encoder--decoder. All models have their largest errors for the undrained test at $e_0 = 0.575$, the densest specimen, which lies outside the training range. \Cref{fig:2_graphs_cycle} compares predictions under drained and undrained tests. The rBNN accurately reproduced cyclic stress--strain behavior, capturing contractive vs.\ dilative responses and stress degradation: the progressive decrease of $p'$ and the degradation of the deviatoric stress amplitude in the undrained test, and the accumulation of volumetric strain in the drained test. The 95\% intervals contain 98--100\% of the simulated values; five of the six test specimens are reliable, the exception being the extrapolated undrained test at $e_0 = 0.575$, whose deviatoric stress is covered at 86\% of the load steps (\cref{tab:uncertainty}; all specimens in Supplementary Fig.~S2).

\subsection{Evaluation on experimental data} \label{sec:Experimental_Data}
\subsubsection{Monotonic Loading}\label{sec:Experimental_Data_mono}

The laboratory dataset comprises 28 CD triaxial tests on Baskarp sand conducted at Aalborg University \cite{ibsen1994baskarp}; two further tests of the series (9301.06 and 9301.07) were excluded because their records are incomplete. Tests span three initial void ratios (\(e_0=0.85,0.70,0.61\)) and ten confining pressures ($\sigma'_3 = 5$--$800$ kPa). Soil samples were subjected to axial strains up to maximum values between 12\% and 19\% of the initial sample height, depending upon the specific test conditions.

The initial void ratios in the data files were checked against the test sheets of the source report \cite{ibsen1994baskarp}. For tests 9301.19--9301.32, the files give $e_0 = 0.852$, the value of the loose series, whereas the sheets give 0.695--0.705 for tests 9301.20--9301.29 (medium-dense series) and 0.614 for test 9301.32 (dense series; Supplementary Table~S1). With the file values, specimens with nominally identical inputs responded clearly differently (Supplementary Fig.~S6). The initial void ratios were corrected and the void-ratio paths rebuilt from the measured volumetric strain, $e_t = e_0 - (1+e_0)\,\varepsilon_{v,t}$; all results reported here use the corrected data.

For training, 16 tests at $\sigma'_3 = 20$--$320$\,kPa were used, while the six tests at the extreme pressures (5 and 800\,kPa) formed the test set and the six tests at 10 and 640\,kPa the validation set (\cref{tab:splits}). Stress invariants ($p',q$) and void ratio ($e$) were normalized following the simulated case (\cref{sec:Data_norm}). Representative stress--strain curves are shown in \cref{fig:datasets}b.

The best configurations used $H = 8$ (rFFNN) and $H = 4$ (rBNN). ReLU and SiLU activations were both admitted for the step models; the recurrent cells use the standard sigmoid and tanh gates with a linear output layer (\cref{tab:hyper}).

\begin{figure}[!ht]
    \centering
    \includegraphics[width=0.92\textwidth]{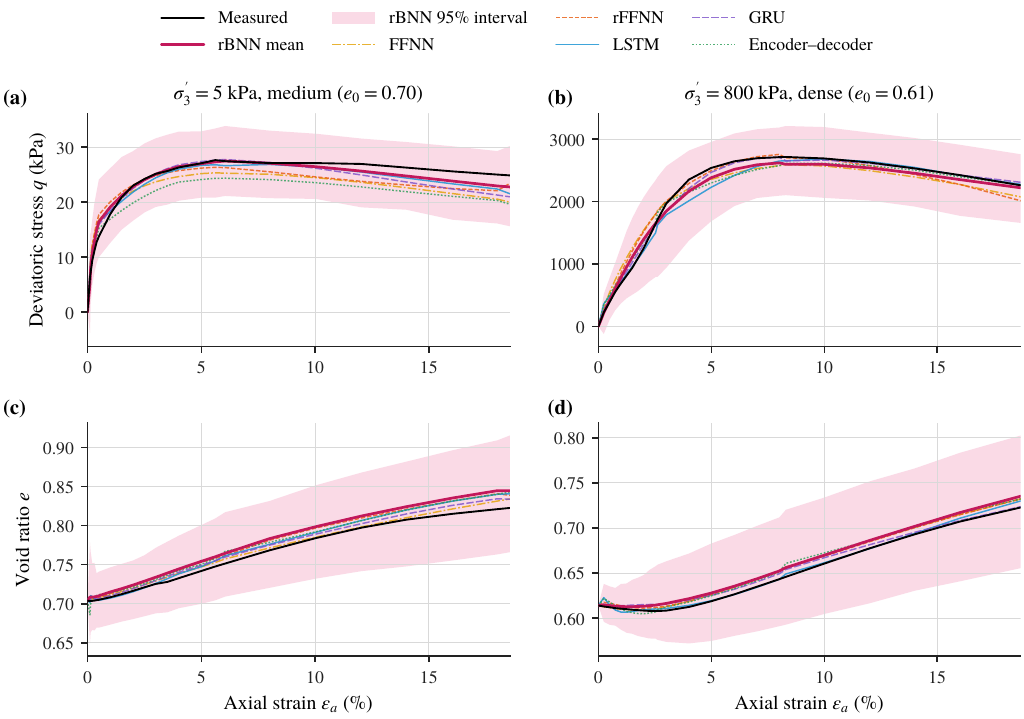}
    \caption{Comparison of model predictions for \textbf{deviatoric stress} (\ensuremath{q}) and \textbf{void ratio} versus \textbf{axial strain} (\ensuremath{\varepsilon}) at two extreme confining pressures in experimental CD tests. The left column shows 5~kPa and the right column shows 800~kPa results for specimens with $e_0 = 0.704$ (test 9301.26) and $e_0 = 0.614$ (test 9301.32), respectively. Each subplot overlays predictions from various ML models, with the rBNN output including 95\% confidence intervals.}
    \label{fig:4_graphs}
\end{figure}

This is the smallest dataset and the most demanding extrapolation: all test specimens lie outside the confining-pressure range of the final training set (10--640\,kPa). The LSTM has the highest mean $R^2$ (0.956), followed by the GRU, the rFFNN and the rBNN (0.946 each); the GRU has the lowest stress errors and the rBNN the second lowest (RMSE of 24.8 and 63.3\,kPa for $p'$ and $q$), while MC dropout predicts the void ratio best (\cref{tab:accuracy}). \Cref{fig:4_graphs} shows a medium-dense specimen at 5\,kPa and a dense specimen at 800\,kPa, one of the corrected tests. The rBNN mean follows the measured deviatoric stress at both pressures, including the post-peak softening at 800\,kPa, and the contraction followed by dilation of the dense specimen; at 5\,kPa, it slightly overestimates the dilation at large strains (void ratio of 0.845 against 0.823 at 18\% strain). The rBNN intervals are considerably wider here (median width 26\% of the data range, against 9\% for the simulated monotonic tests; \cref{tab:uncertainty}). This is in contrast to the simulated dataset, where the larger dataset size (78 tests) allowed the rBNN to produce tighter confidence intervals. The wider intervals reflect the conditions of this dataset: only 22 tests, spread over three densities and eight pressure levels, are available for the final models; every test specimen requires extrapolation in $\sigma'_3$; and the volume-change measurements scatter at large strains. The intervals are on the cautious side, i.e., wider than necessary at intermediate nominal levels (Supplementary Fig.~S5), and at the 95\% level they contain 95--100\% of the measured values. Five of the six test specimens are reliable. The exception is the dense specimen at 5\,kPa ($e_0 = 0.616$; Supplementary Fig.~S3), whose high peak stress ratio and pronounced softening at very low confinement are not represented in the training data; it has the largest errors for all models and lies outside the rBNN interval at 35\% of the load steps.

\subsubsection{Cyclic Loading}\label{sec:Experimental_Data_cycle}
The cyclic loading dataset was taken from \cite{elghoraiby2020stress}, consisting of 35 undrained strain-controlled triaxial tests on Ottawa F65 sand. Specimens were prepared using the constant height dry pluviation method and consolidated to initial void ratios of 0.576, 0.608, and 0.668, corresponding to relative densities of 75.6\%, 63.8\%, and 40.0\%. Within each group, the cyclic axial strain amplitude (CSA) was varied, from 0.040\% to 0.140\% for $e_0 = 0.668$ (9 tests), 0.050\% to 0.220\% for $e_0 = 0.608$ (14 tests), and 0.050\% to 0.300\% for $e_0 = 0.576$ (12 tests) to investigate liquefaction resistance. Liquefaction triggering was defined by reaching an excess pore pressure ratio $r_u=0.95$, with $r_u = (u - u_0)/\sigma'_{3,0}$, where $u_0$ is the pore pressure and $\sigma'_{3,0}$ the effective confining stress at the start of cycling.

Data preprocessing followed the same normalization strategy as in \cref{sec:Data_norm}. From each void-ratio group, one test was held out for testing (CSA $= 0.25\%$, 0.22\% and 0.14\% for $e_0 = 0.576$, 0.608 and 0.668: the largest amplitude of the two looser groups and the second largest of the densest) and one for validation (CSA $= 0.15\%$, 0.125\% and 0.10\%), leaving 29 training specimens (\cref{tab:splits}). 

The best configurations used $H = 13$ for both the rFFNN and the rBNN (\cref{tab:hyper}), again longer windows than for monotonic loading.

\begin{figure}[!ht]
    \centering
    \includegraphics[width=0.92\textwidth]{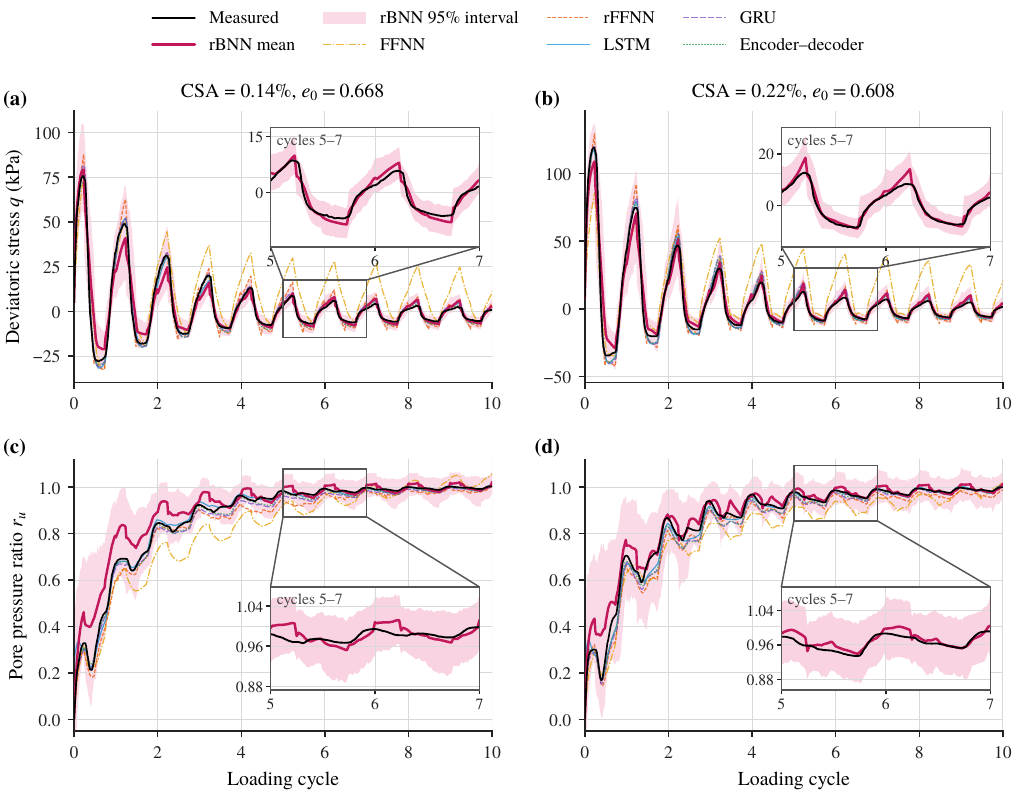}
    \caption{Comparison of model predictions for \textbf{deviatoric stress} $q$ (\textbf{a}, \textbf{b}) and \textbf{excess pore pressure ratio} $r_u$ (\textbf{c}, \textbf{d}) versus loading cycle from experimental cyclic undrained tests. Left column shows results for CSA = 0.14\% and initial void ratio $e_0 = 0.668$, and right column for CSA = 0.22\% and $e_0 = 0.608$. Each subplot includes predictions from multiple ML models, with the rBNN output displaying 95\% confidence intervals; the insets enlarge cycles 5--7.}
    \label{fig:4_graphs_cycle}
\end{figure}

This dataset shows most clearly why recursive training matters: the FFNN becomes unstable over the roughly 1000 recursive steps of a test (mean $R^2 = -6.6$), whereas the rFFNN, which differs only in the training window, reaches 0.956 (\cref{tab:accuracy}). The recurrent models are the most accurate (0.980--0.983); the rBNN (0.953; RMSE of 3.0\,kPa, 4.5\,kPa and 0.034 for $p'$, $q$ and $r_u$) is close to the rFFNN and the deep ensemble. MC dropout fails here (mean $R^2 = 0.05$): with one dropout mask per trajectory, its trajectories diverge from one another, so the pooled mean is poor and the interval spans the whole data range. \Cref{fig:4_graphs_cycle} shows that the rBNN reproduces the degradation of the deviatoric stress amplitude and the build-up of $r_u$ towards unity. For the loosest specimen, it overestimates $r_u$ in the first three cycles, but the measured response stays inside the interval (\cref{fig:4_graphs_cycle}c). The 95\% intervals contain 98--100\% of the measured values, all three test specimens are reliable, and the intervals are the narrowest of the three probabilistic models (median width 11\% of the data range, \cref{tab:uncertainty}; all specimens in Supplementary Fig.~S4). The repeated loading paths of the cyclic records provide richer training signals than the short monotonic records, consistent with the narrower intervals than for Baskarp sand. This makes the rBNN suitable for capturing complex behaviors like cyclic softening and pore pressure buildup under undrained conditions.

\begin{figure}[!ht]
    \centering
    \includegraphics[width=0.92\textwidth]{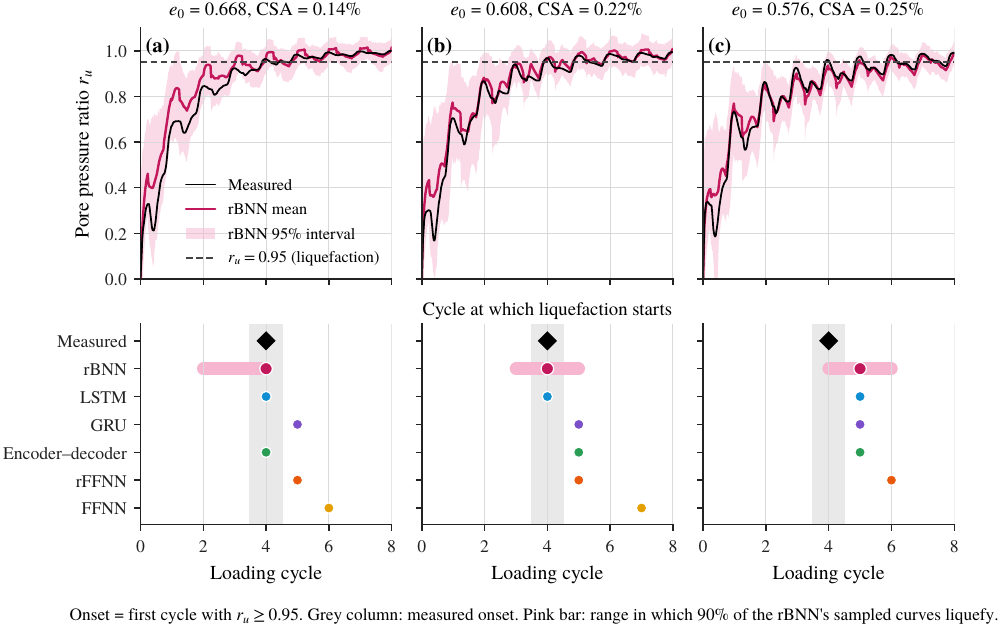}
    \caption{\textbf{Onset of liquefaction} for the three Ottawa F65 test specimens. Top: measured excess pore pressure ratio, rBNN mean and 95\% interval, and the threshold $r_u = 0.95$. Bottom: cycle in which $r_u$ first reaches 0.95 (cycles counted from the zero crossings of the axial strain), measured (diamond, grey column) and predicted by each model; the pink bar spans the 5--95\% range of the onset cycles of the 200 rBNN trajectories (weights fixed along each trajectory; all reached the threshold).}
    \label{fig:liquefaction}
\end{figure}

As an engineering measure, \cref{fig:liquefaction} compares the predicted and measured onset of liquefaction, a quantity of the kind targeted by cycle-based models \cite{jas2025evaluation,jas2026comparative}. The onset is the first load step at which $r_u = (u-u_0)/\sigma'_{3,0} \ge 0.95$. Every load step is assigned to a loading cycle from the zero crossings of the axial strain, and the onset is given both as the number of cycles completed and as the cycle in which it occurs. The measured $r_u$ first reaches 0.95 after 3.88, 3.90 and 3.91 cycles ($e_0 = 0.668$, 0.608 and 0.576), i.e., in the fourth cycle for all three test specimens. El Ghoraiby et al.\ \cite{elghoraiby2020stress} report 3.89, 3.90 and 4.89 cycles for these tests. Their values are reproduced from the same records (3.89, 3.93 and 4.89 cycles) when liquefaction is defined by a 95\% reduction of the mean effective stress, $1 - p'/p'_0 \ge 0.95$: for the densest specimen, the deviatoric stress is still large when $r_u$ reaches 0.95, so $p'$ falls to 5\% of $p'_0$ only one cycle later.

For the rBNN, the onset is evaluated for every sampled trajectory: each of the 100 weight samples of each retained configuration (two for this dataset, i.e., 200 trajectories) is held fixed over the record, its predicted mean is fed back (\cref{sec:predictive}), and the onset is read from the predicted-mean path of $r_u$ without adding the aleatoric noise. Trajectories that never reach the threshold within the record are excluded from the onset range, and their fraction is reported; for the three test specimens, all rBNN trajectories reached it. \Cref{tab:onset} lists the results. The rBNN mean predicts the fourth cycle for the two looser specimens and the fifth for the densest; 90\% of the sampled trajectories liquefy within cycles 2--4, 3--5 and 4--6, which contain the measured cycle for all three specimens. The LSTM predicts the same cycles as the rBNN mean, the GRU and the rFFNN are one to two cycles late, the encoder--decoder up to one cycle late, and the FFNN two to six cycles late.

{
\begin{table}[!ht]

\footnotesize
\caption{\textbf{Onset of liquefaction} ($r_u \ge 0.95$) for the three Ottawa F65 test specimens. Measured onset in cycles completed and as the cycle in which it occurs; values reported by El Ghoraiby et al.\ \cite{elghoraiby2020stress}, who define liquefaction differently (see text); rBNN onset cycle of the predictive mean, and median and 5--95\% range over the 200 sampled trajectories, all of which reached the threshold.}
\label{tab:onset}
\begin{tabularx}{\textwidth}{>{\raggedright\arraybackslash\hsize=1.5\hsize}X>{\centering\arraybackslash\hsize=0.9\hsize}X>{\centering\arraybackslash\hsize=0.9\hsize}X>{\centering\arraybackslash\hsize=0.9\hsize}X>{\centering\arraybackslash\hsize=0.9\hsize}X>{\centering\arraybackslash\hsize=0.9\hsize}X}
\toprule
\textbf{Test specimen} & \textbf{Measured, cycles completed} & \textbf{El Ghoraiby et al.} & \textbf{Measured cycle} & \textbf{rBNN mean, cycle} & \textbf{rBNN median (5--95\%)} \\
\midrule
$e_0 = 0.668$, CSA $= 0.14\%$ & 3.88 & 3.89 & 4 & 4 & 3 (2--4) \\
$e_0 = 0.608$, CSA $= 0.22\%$ & 3.90 & 3.90 & 4 & 4 & 4 (3--5) \\
$e_0 = 0.576$, CSA $= 0.25\%$ & 3.91 & 4.89 & 4 & 5 & 5 (4--6) \\
\bottomrule
\end{tabularx}
\end{table}}

\subsection{\texorpdfstring{Reliability of the predictive intervals}{Reliability of the predictive intervals}}\label{sec:calibration}
\Cref{tab:uncertainty} compares the three probabilistic models on every test specimen. The rBNN intervals contain 95--100\% of the observed values of every state variable on all four datasets, and 19 of the 21 test specimens are reliable; the exceptions are the two extrapolated specimens discussed above (simulated undrained test at $e_0 = 0.575$, 86\% coverage; dense Baskarp specimen at 5\,kPa, 65\%). The deep ensemble gives the narrowest intervals and the best interval score on the simulated datasets, but these intervals are over-confident: 13 of the 21 specimens are reliable, its reliability diagrams lie below the diagonal on three datasets (Supplementary Fig.~S5), and on Baskarp sand it covers only 80--82\% of the measured stresses (worst specimen 22\%). MC dropout covers 20 of the 21 specimens, but its intervals are 1.3--3.7 times wider than those of the rBNN on the simulated datasets and span the whole data range on the Ottawa F65 tests, where its mean prediction is poor. On Baskarp sand, MC dropout and the rBNN perform similarly (interval scores of 0.29 and 0.31). Overall, the rBNN offers the most consistent balance of accuracy, coverage and interval width across the four datasets, without being the best model on every criterion.

All three probabilistic models are pooled over their retained configurations. \textit{Single} networks differ more (Supplementary Table~S3): the median coverage of a single rBNN is 0.90, 0.97, 0.90 and 0.97 on the four datasets, against 0.51, 0.67, 0.53 and 0.94 for a single network with a variance output (one deep-ensemble member) and 0.94--1.00 for single MC dropout networks. Because the deep ensemble and the rBNN share architecture and training windows, this isolates the effect of the Bayesian weights: they make the intervals of an individual network substantially more reliable, so the reliability of the rBNN does not hinge on pooling, while pooling improves all three models.

The rBNN intervals are generalized-Bayesian intervals (\cref{sec:rBNN}), calibrated empirically rather than by construction. Their reliability diagrams lie above the diagonal for the simulated cyclic and Baskarp datasets (intervals wider than necessary at intermediate levels) and slightly below it for the simulated monotonic stresses at nominal levels of 0.6--0.9 (largest shortfall 0.12; Supplementary Fig.~S5). Where calibration at all levels matters, the interval widths could be rescaled post hoc on the validation tests without retraining.

\input{tables/uncertainty}

\subsection{\texorpdfstring{Ablation study}{Ablation study}}\label{sec:ablation}
\Cref{tab:ablation} changes one ingredient of the rBNN at a time; every variant is tuned, retrained and pooled like the main models. \textit{Recursive training} (A1 to A2) is the key ingredient for accuracy: it raises the mean $R^2$ from 0.892 to 0.996 for the simulated cyclic tests and from $-6.6$ to 0.956 for the Ottawa F65 tests, with moderate gains on the monotonic datasets. \textit{Bayesian weights without recursive training} (A4) do not replace it: the one-step rBNN is well covered but inaccurate under cyclic loading ($R^2 = 0.385$ and $-1.1$). \textit{A variance output alone} (A3, the deep ensemble) keeps the accuracy of the rFFNN but is over-confident on three datasets (lowest PICP of 0.80--0.93), whereas \textit{the combination} (A5, rBNN) keeps the accuracy and reaches a lowest coverage of 0.95--0.98.

R\'enyi divergences \cite{li2016renyi} with $\alpha = 0.5$ or $2$ (A6, A7) change neither accuracy nor coverage consistently, so the KL divergence is retained. Down-weighting the KL term by $10^{-4}$ (A8) slightly improves the simulated monotonic fit but lowers accuracy and coverage on both experimental datasets (mean $R^2$ of 0.933 and 0.907, lowest coverage 0.92 and 0.93): the prior regularization matters most where data are scarce and noisy. A new weight sample at every load step (A9) leaves the mean prediction unchanged but lowers the coverage, most clearly for the simulated monotonic tests (0.93 instead of 0.98), because it captures only one-step-ahead uncertainty. Feeding back a sampled instead of the mean state (A10) gives similar coverage and a slightly lower accuracy on the Ottawa F65 tests; the mean feedback of \cref{eq:joint_likelihood} is retained.

\input{tables/ablation}

\subsection{\texorpdfstring{Physical consistency of the recursive predictions}{Physical consistency of the recursive predictions}}\label{sec:physics}
No physical constraints (yield or critical-state conditions, thermodynamic admissibility, bounds on $r_u$) are imposed in training or prediction; apart from the exact relation $p' = \sigma'_3 + q/3$ used for the simulated drained monotonic tests, the rBNN is purely data-driven. Physical consistency is promoted by the formulation: (i) the network maps the current \textit{physical} state and the prescribed strain input to the next state, so that path dependence is carried by measurable state variables rather than by latent memory; (ii) predictions are conditioned on $\ve{\theta}$; and (iii) the loading-direction indicator $\Delta_t$ allows different stiffness on loading and unloading. The following checks whether the predictions honour the key mechanisms of sand behavior.

\textit{Dilatancy and stress ratio.} Under drained monotonic loading, the rBNN reproduces the contraction followed by dilation of the dense Baskarp specimen at 800\,kPa and the dilative response at 5\,kPa (\cref{fig:4_graphs}), as well as the corresponding trends of the simulated tests (\cref{fig:1_graphs}). The mean absolute error of the predicted peak stress ratio $\eta_{\max} = \max_t (q/p')$ is 0.017 for the simulated and 0.047 for the Baskarp test specimens (LSTM: 0.005 and 0.040). In the simulated cyclic tests, the contractive accumulation of volumetric strain under drained loading and the associated stiffness degradation are reproduced (\cref{fig:2_graphs_cycle}).

\textit{Pore pressure accumulation and cyclic softening.} Under undrained cyclic loading, the predicted mean effective stress decreases progressively and the deviatoric stress amplitude degrades as $r_u$ accumulates towards unity (\cref{fig:2_graphs_cycle,fig:4_graphs_cycle}), consistent with cyclic softening and liquefaction, and the measured onset of liquefaction lies within the predicted range (\cref{fig:liquefaction}).

\textit{Admissibility bounds.} The predictive mean of $p'$ is positive at every load step of every test specimen; among the sampled rBNN trajectories, negative values occur at 0.05\% of the load steps of the cyclic test specimens (MC dropout: 1.3--1.6\%). The excess pore pressure ratio is not bounded: the measured $r_u$ reaches 1.007, the rBNN mean up to 1.04, and the upper edge of the 95\% interval exceeds unity over 61--84\% of the load steps of the three test specimens, by up to 0.10 (\cref{fig:4_graphs_cycle}c,d); such values should be truncated at $r_u = 1$ in practice. Bounded output transformations and physics-based constraints (e.g., critical-state or stress--dilatancy relations) in the loss are left for future work (\cref{sec:discussion}).

\subsection{Computational time}\label{sec:time}
All models were implemented in one PyTorch code base (Python 3.10, PyTorch 2.5.1 with CUDA 12.1) and trained on one NVIDIA RTX A6000 GPU, which was shared with other jobs; the times in \cref{tab:cost} are therefore indicative and best compared within a dataset. 

One final model trains in 0.3--9\,min, except the rBNN on the simulated cyclic dataset (12.5\,min; \cref{tab:cost}). The FFNN is the fastest model and the recurrent models train in a few minutes; on the experimental datasets, the rBNN is comparable to the other models. Its higher cost stems from the doubled number of variational parameters, the sampling of one set of weights per training window and the slower convergence of the KL-regularized objective.

All models received the same number of search configurations, but not the same computing time. Because an rBNN configuration is more expensive to train, the rBNN has the largest search time on three of the four datasets and in total (401\,min, against 130--302\,min for the recurrent models and 216--237\,min for the two uncertainty baselines). To check whether the remaining accuracy gap to the LSTM is a matter of search budget, the rBNN search on the simulated monotonic dataset was repeated with 24 instead of 12 configurations: the best validation score improved, but the test accuracy did not (mean $R^2$ of 0.987 against 0.992) and the coverage decreased, so the gap is not closed by more tuning at this scale. At prediction time, the rBNN needs one recursive pass per weight sample (100 per retained configuration), computed in parallel.

\input{tables/cost}

Overall, the LSTM offers the best accuracy at a moderate cost, the FFNN and rFFNN are the cheapest models, and the rBNN trades a higher training cost for reliable uncertainty quantification---valuable when reliability is critical.

\section{Discussion} \label{sec:discussion}

This study highlights the potential of recursive neural network architectures, particularly rFFNNs and rBNNs, in modeling the constitutive behavior of sand under both monotonic and cyclic loading conditions. Unlike recurrent models such as LSTMs and GRUs, which rely on hidden states and gating mechanisms, rFFNNs provide an explicit state-to-state mapping, expressed entirely in measurable state variables, between input features and output states across load steps. This explicit structure makes it easier to analyze the impact of specific input features on the predicted state variables, making rFFNNs and their Bayesian counterpart (rBNNs) valuable tools in geotechnical modeling.

One distinguishing feature of rFFNNs and rBNNs is the use of a sliding window length $H$, which defines the temporal context for training and recursive predictions. Unlike LSTMs or GRUs, which inherently model sequential dependencies through hidden states that evolve across the full loading sequence, rFFNNs and rBNNs require an explicit window length. The choice of $H$ plays a significant role in predictive performance; in particular, the comparison of the FFNN ($H = 1$) and the rFFNN shows that training over recursive windows is what makes long recursive predictions stable under cyclic loading. The selected window lengths depend on the type of loading: $H = 4$--$9$ for the monotonic and $H = 13$--$16$ for the cyclic datasets, consistent with the longer context needed to represent repeated loading and unloading.

From a performance perspective, the recurrent baselines, in particular the LSTM, remain slightly more accurate than the rBNN when all models are tuned with the same search budget; the differences in mean $R^2$ are 0.003--0.03. Additionally, the rBNN provides an essential advantage in uncertainty quantification, producing confidence intervals that were reliable on 19 of the 21 held-out test specimens, more consistently than a deep ensemble of the same architecture or MC dropout. This feature is particularly beneficial in geotechnical applications where predictive confidence is critical for risk assessment: the predicted onset of liquefaction, for example, is given as a range of cycles that contained the measured onset for all three Ottawa F65 test specimens. These benefits come at the highest training cost of the models compared: with an equal number of tuning configurations, the rBNN required the largest total search time, and doubling its search budget did not improve its test accuracy (\cref{sec:time}).

On experimental datasets, the wider confidence intervals observed in monotonic experimental data highlight the challenges posed by data sparsity and measurement noise, as opposed to cyclic experimental datasets where enough data is available. This suggests that the rBNN's probabilistic framework, while robust, requires careful calibration and potentially enhanced priors or noise models for sparse experimental data. The correction of the Baskarp void ratios (\cref{sec:Experimental_Data_mono}) also illustrates that, for small laboratory datasets, the consistency of the test metadata deserves the same scrutiny as the model.

\textit{Limitations.} Several limitations of the present study should be noted. (i) Each configuration was trained once with a fixed seed; pooling the retained configurations reduces, but does not remove, the effect of run-to-run variability, and the small differences in accuracy between the best models (\cref{tab:accuracy}) should not be over-interpreted. (ii) Generalization was assessed within each sand, to new confining pressures, strain amplitudes and initial void ratios, not across sands or loading paths such as simple shear, and extrapolation of the stress level remains difficult for all models compared (\cref{sec:Simulation_Data_mono}). (iii) The models are dataset-specific and predict responses to prescribed strain-controlled triaxial histories; they are not formulated as stress-point constitutive laws for arbitrary loading paths and have not been implemented in a boundary-value (e.g., finite-element) solver.

\section{Conclusion} \label{sec:concusion}
This study presents a novel uncertainty-aware recursive Bayesian neural network (rBNN) framework for modeling the stress-strain-volume response of sands under monotonic and cyclic triaxial loading, as dataset-specific predictors of the response to prescribed strain histories. By enhancing recursive feedforward neural networks (rFFNNs) through a sliding-window training strategy and embedding them within a generalized Bayesian inference framework, the model achieves competitive predictive performance together with quantified predictive uncertainty. The following conclusions are drawn:

The proposed sliding window approach stabilizes training by limiting recursive depth and facilitating localized error correction, enabling better learning of long-range temporal dependencies under cyclic loads. It is the decisive ingredient for long recursive predictions: training over recursive windows raised the mean $R^2$ of the step model from $-6.6$ to 0.956 for the experimental cyclic tests. The Bayesian formulation, built upon a deterministic Gaussian output layer and generalized variational inference, introduces principled uncertainty quantification in model predictions while mitigating overfitting.

Comparative evaluations on both simulated and experimental datasets show that the rBNN performs competitively with state-of-the-art deterministic models such as LSTM, GRU, and Encoder--Decoder architectures, all tuned with an equal search budget. Its accuracy is close to, but not higher than, that of the best recurrent model (mean $R^2$ of 0.95--0.99, within 0.003--0.03 of the LSTM). Its 95\% predictive intervals contained at least 90\% of the measured response on 19 of the 21 test specimens, against 13 of 21 for a deep ensemble of the same architecture, while being narrower than those of MC dropout on three of the four datasets; for the experimental cyclic tests, the measured onset of liquefaction lay within the predicted range for all test specimens. These benefits come at the highest training cost of the models compared.

This work establishes a probabilistic recursive modeling approach for the constitutive modeling of sands, setting the foundation for future research into fully Bayesian sequence models and their applications in reliability analysis and field-scale monitoring of geomaterials.

\section*{Acknowledgements}
\textbf{T. Noor} acknowledges the financial support received from the Prime Minister's Research Fellowship (PMRF) and \textbf{R. Nayek} acknowledges the financial support received from ANRF (vide grant no.\ SRG/2022/001410), Ministry of Shipping, Ports, and Waterways, and the matching grant from IIT Delhi.

\section*{Author contributions}
The authors confirm contribution to the paper as follows: Conceptualization, Toiba Noor and Rajdip Nayek; methodology, Toiba Noor, Soban Nasir Lone and Rajdip Nayek; software, Toiba Noor; writing---original draft preparation, Toiba Noor; writing---review and editing, G. V. Ramana and Rajdip Nayek; supervision, G. V. Ramana and Rajdip Nayek. All authors reviewed and approved the final version of the manuscript.

\section*{Data availability}
Some or all data, models, or code that support the findings of this study are available from the corresponding author upon reasonable request.

\section*{Ethics approval}
Not applicable.

\section*{Declaration of competing interest}
The authors declare no conflicts of interest.

\section*{Supplementary material}
The supplementary material is provided at the end of this document. Figures S1--S4: rBNN predictions and 95\% intervals for every test specimen of the four datasets; Figure S5: reliability diagrams; Figure S6: Baskarp sand responses before and after the correction of the initial void ratios; Figure S7: validation score versus window length; Figure S8: training time per model; Figure S9: coverage of every test specimen; Table S1: corrected initial void ratios of the Baskarp tests; Table S2: configurations retained in the pooled predictions; Table S3: coverage of single networks.

\section*{Abbreviations}
\indent The following abbreviations are used in this manuscript:

\noindent
{
\begin{tabular}{@{}ll}
BNN & Bayesian neural network\\
CD / CU & Consolidated drained / consolidated undrained\\
CRPS & Continuous ranked probability score\\
CSA & Cyclic axial strain amplitude\\
ELBO & Evidence lower bound\\
FFNN & Feedforward neural network\\
GRU & Gated recurrent unit\\
GVI & Generalized variational inference\\
IS & Interval score\\
KL & Kullback--Leibler\\
LSTM & Long short-term memory\\
MAE & Mean absolute error\\
MC & Monte Carlo\\
NLL & Negative log-likelihood\\
PICP & Prediction interval coverage probability\\
PINAW & Prediction interval normalized average width\\
rBNN & Recursive Bayesian neural network\\
rFFNN & Recursive feedforward neural network\\
RMSE & Root mean squared error\\
RNN & Recurrent neural network\\
UQ & Uncertainty quantification\\
VI & Variational inference
\end{tabular}}

\appendix
\section{Training and Optimization of RNNs}\label{appendix:implementation}
The data normalization process for RNNs mirrors the approach used for rFFNN and rBNN. The RNN models were trained using input data structured as shown in \cref{fig:sample_ffnn_single}. RNNs maintain an internal \textit{hidden state} $\ve{h}_t^{(m)}$ that acts as a memory of past inputs. For each instance $m$ and at each time step $t$, the current hidden state is computed as a function of the previous hidden state, $\ve{h}_{t-1}^{(m)}$, the current input, $\mathbf{u}_t^{(m)}$, the triaxial test constants $\ve{\theta}^{(m)}$, and the model's trainable parameters $\ve{\psi}$, and the state is read out from the hidden state. This feedback mechanism is described by:
\input{figure2}

{
\begin{align}
\ve{h}_t^{(m)} = f_{\ve{\psi}}\left(\ve{h}_{t-1}^{(m)}, \mathbf{u}_t^{(m)}, \ve{\theta}^{(m)}\right), \qquad \hat{\ve{s}}_t^{(m)} = g_{\ve{\psi}}\left(\ve{h}_{t}^{(m)}\right)
\end{align}}
where the measured initial state $\ve{s}_0^{(m)}$ is concatenated with $\ve{u}_t^{(m)}$ and $\ve{\theta}^{(m)}$ at every step, so that it remains available throughout the sequence, and the hidden state is initialized with zeros. In the encoder--decoder, the bidirectional encoder processes the complete prescribed input sequence, and the unidirectional decoder produces the states causally from the encoded sequence.

During training, the RNNs process the training records, or random segments of equal length when the records differ in length (\cref{sec:protocol}), incrementally. At each strain increment, the hidden state from the preceding time step is propagated forward and integrated with the current input to update the model's internal memory. Losses are then calculated for the predictions made at each of these increments.

\noindent{}
The RNNs were tuned, trained and pooled with the same protocol and search budget as all other models (\cref{sec:protocol}), with the squared loss. The recurrent cells use the standard sigmoid and tanh gate activations, followed by a linear output layer. The selected configurations are given in \cref{tab:hyper} and Supplementary Table~S2.

\bibliographystyle{unsrtnat}
\bibliography{ref,rev_refs}

\clearpage
\setcounter{figure}{0}\setcounter{table}{0}\setcounter{equation}{0}
\renewcommand{\thefigure}{S\arabic{figure}}\renewcommand{\thetable}{S\arabic{table}}\renewcommand{\theequation}{S\arabic{equation}}
\renewcommand{\theHfigure}{S\arabic{figure}}\renewcommand{\theHtable}{S\arabic{table}}\renewcommand{\theHequation}{S\arabic{equation}}
\section*{Supplementary Material}
\addcontentsline{toc}{section}{Supplementary Material}
{
\noindent This supplement supports \cref{sec:Data_norm,sec:assess} of the main text. All results were obtained with the protocol of \cref{sec:protocol} of the main text: every model was tuned with 12 random configurations, the retained configurations were retrained on the training and validation tests, and their predictions were pooled. ``Window length $H$'' denotes the number of recursive predictions per training window, so that $H = 1$ corresponds to one-step training (FFNN).

\subsection*{S1. Predictions for every test specimen}

Figures~\ref{fig:S1}--\ref{fig:S4} show the rBNN predictive mean and 95\% interval for every test specimen of the four datasets; the main text shows two specimens per dataset. The coverage of every specimen is summarized in Fig.~\ref{fig:S9}.

\begin{figure}[H]
\centering
\includegraphics[width=0.95\textwidth]{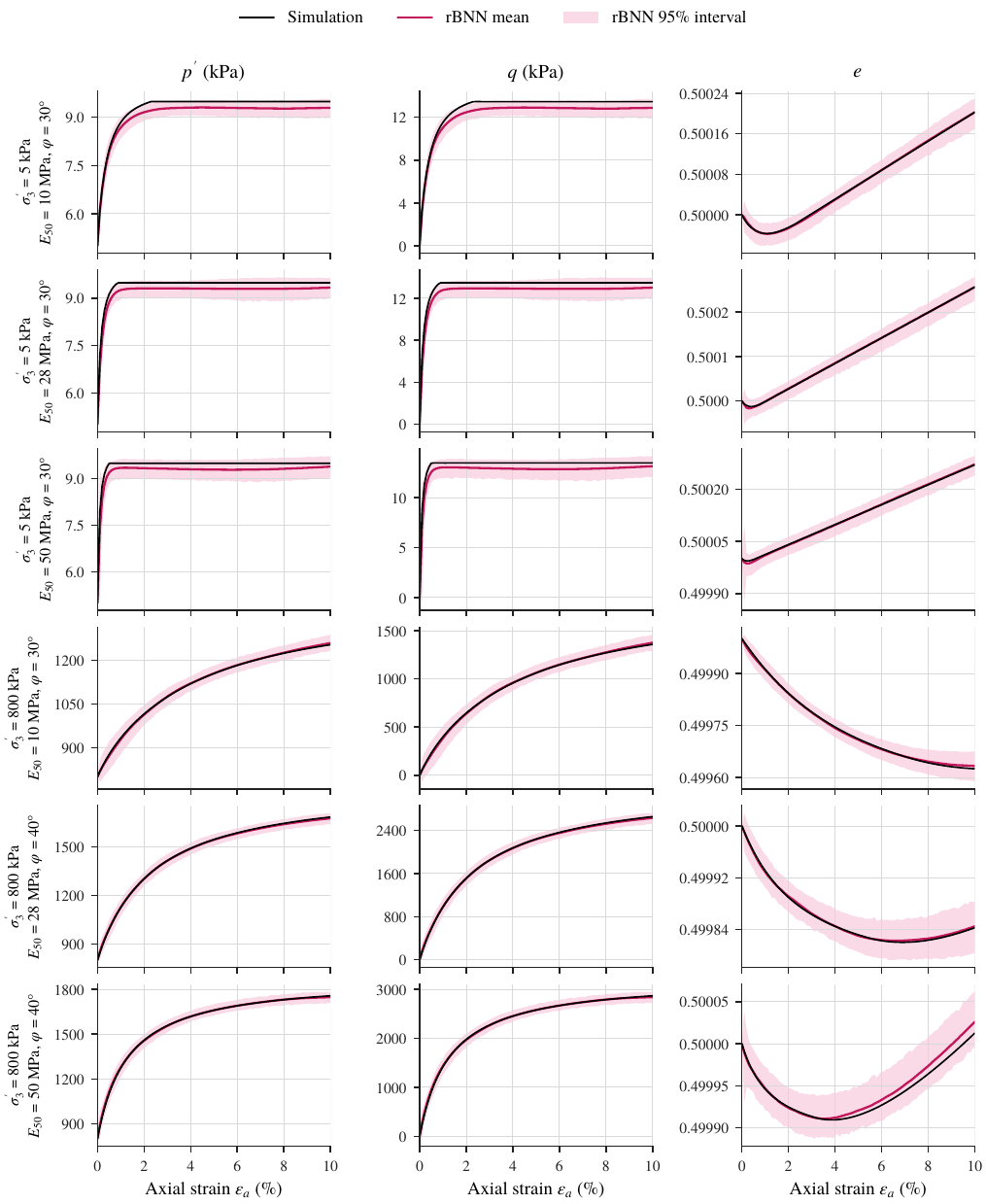}
\caption{Simulated monotonic drained tests: rBNN predictive mean and 95\% interval for the six test specimens (rows), with the mean effective stress $p'$, deviatoric stress $q$ and void ratio $e$ in the columns. $p'$ is obtained from the predicted $q$ as $p' = \sigma'_3 + q/3$.}
\label{fig:S1}
\end{figure}

\begin{figure}[H]
\centering
\includegraphics[width=0.95\textwidth]{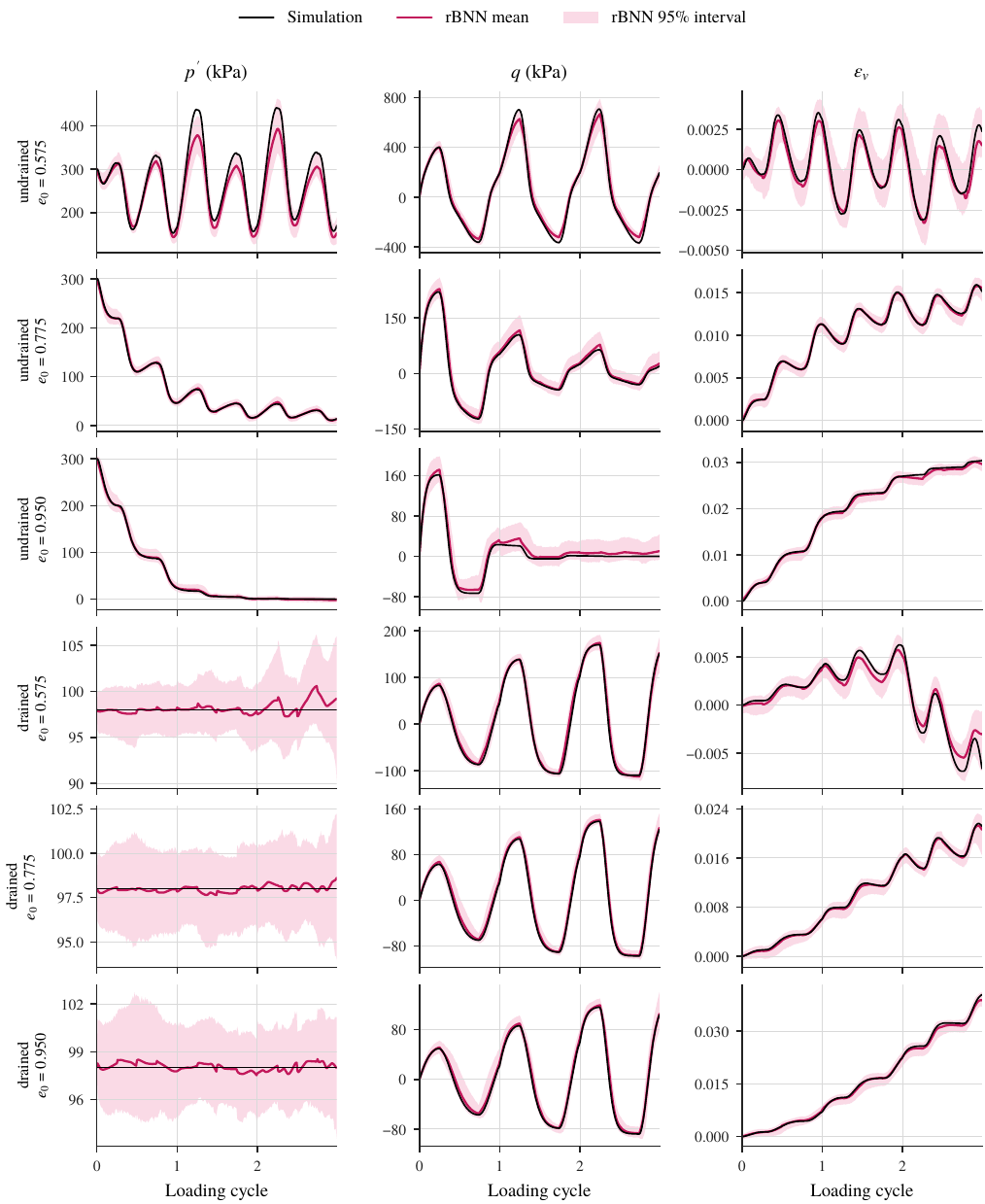}
\caption{Simulated cyclic tests: rBNN predictive mean and 95\% interval for the six test specimens ($e_0 = 0.575$, 0.775 and 0.950; undrained and drained), with $p'$, $q$ and the volumetric strain $\varepsilon_v$ versus loading cycle.}
\label{fig:S2}
\end{figure}

\begin{figure}[H]
\centering
\includegraphics[width=0.95\textwidth]{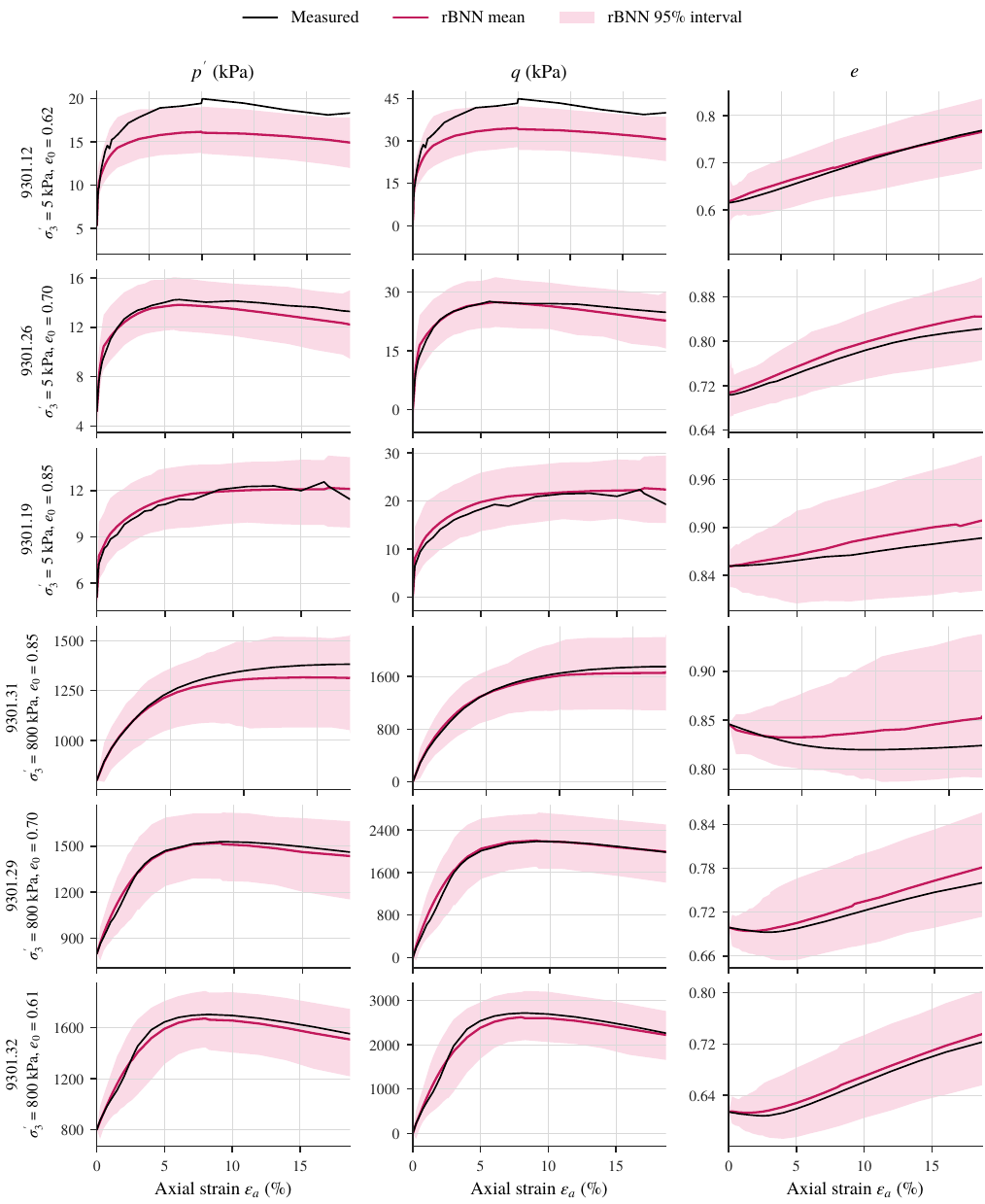}
\caption{Baskarp sand: rBNN predictive mean and 95\% interval for the six test specimens at 5 and 800\,kPa (rows; test number, confining pressure and corrected initial void ratio on the left). The dense specimen at 5\,kPa (9301.12, first row) is the only one whose measured stresses lie partly outside the interval.}
\label{fig:S3}
\end{figure}

\begin{figure}[H]
\centering
\includegraphics[width=0.95\textwidth]{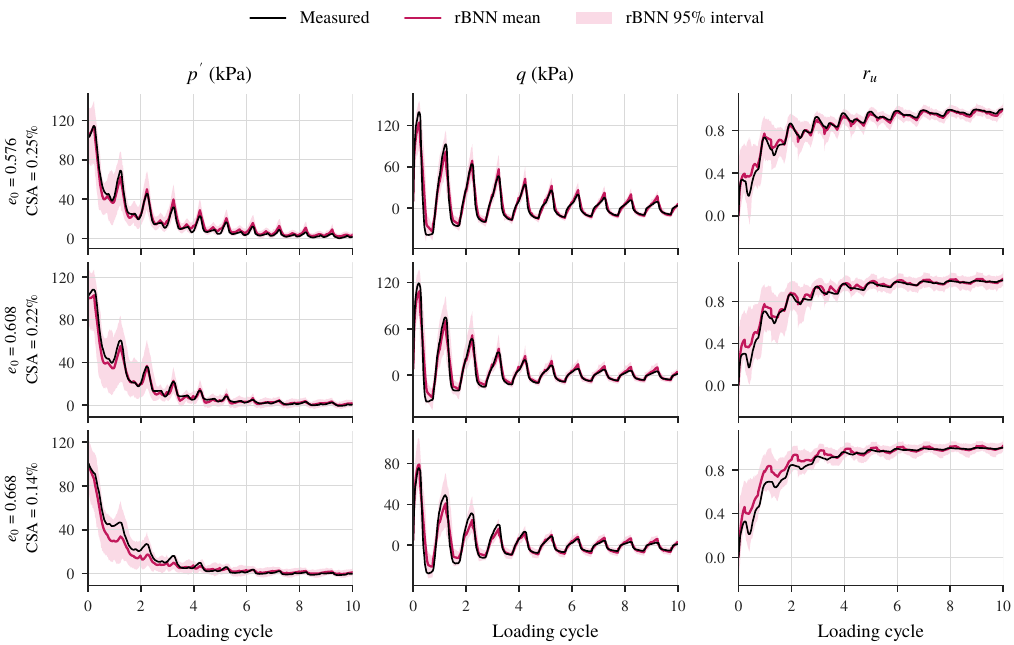}
\caption{Ottawa F65 sand: rBNN predictive mean and 95\% interval for the three test specimens, with $p'$, $q$ and the excess pore pressure ratio $r_u$ versus loading cycle.}
\label{fig:S4}
\end{figure}

\subsection*{S2. Reliability of the predictive intervals}

Figure~\ref{fig:S5} plots, for each probabilistic model, the achieved coverage against the nominal (promised) coverage for central intervals with nominal levels from 0.05 to 0.95, taking for each nominal level the worst of the three state variables. Curves above the diagonal indicate intervals that are wider than necessary (cautious); curves below the diagonal indicate over-confident intervals. The ``worst shortfall'' is the largest distance below the diagonal. Figure~\ref{fig:S9} shows the coverage of the 95\% interval for every test specimen.

\begin{figure}[H]
\centering
\includegraphics[width=\textwidth]{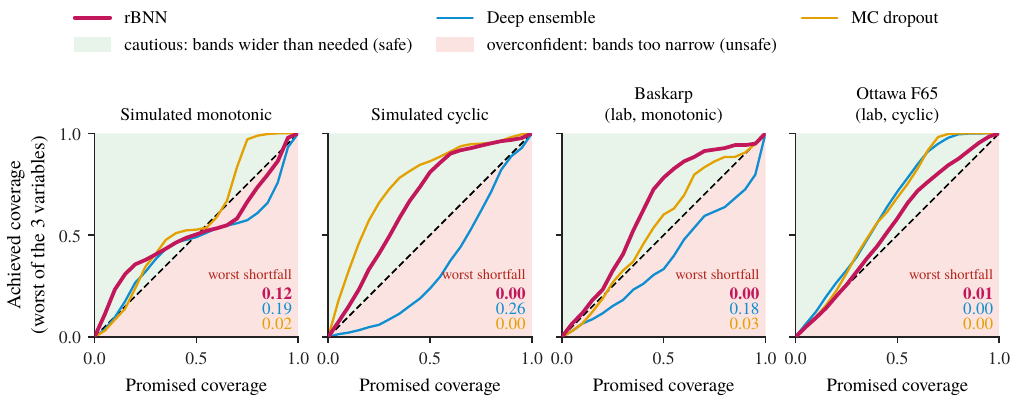}
\caption{Reliability diagrams of the rBNN, the deep ensemble and MC dropout on the test sets of the four datasets (worst of the three state variables at each nominal level). Green: cautious region; red: over-confident region.}
\label{fig:S5}
\end{figure}

\setcounter{figure}{8}
\begin{figure}[H]
\centering
\includegraphics[width=\textwidth]{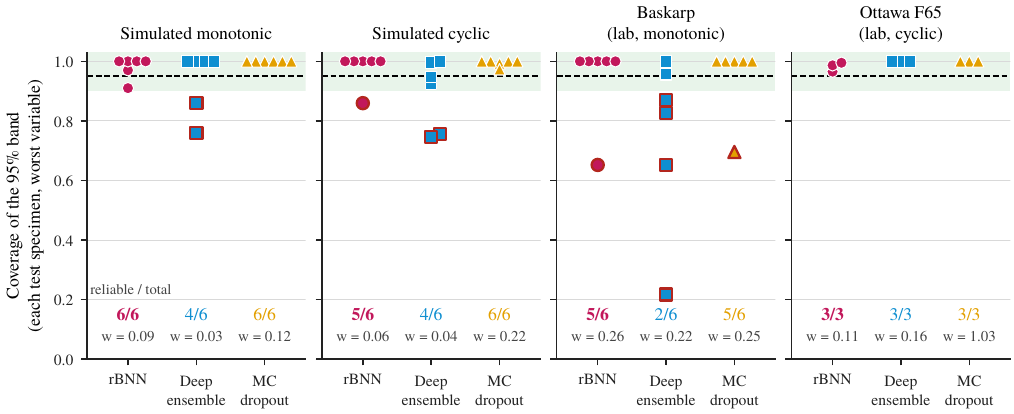}
\caption{Coverage of the 95\% interval for every test specimen (worst of the three state variables). The shaded band marks coverages of at least 0.90 (``reliable'' specimens), the dashed line the nominal level of 0.95; outlined markers are specimens below 0.90. Below each model: number of reliable specimens and median interval width $w$ relative to the data range.}
\label{fig:S9}
\end{figure}
\setcounter{figure}{5}

\setcounter{table}{2}
\input{tables/S3_single}

\subsection*{S3. Correction of the initial void ratios of the Baskarp tests}

For tests 9301.19--9301.32, the digital data files give the initial void ratio of the loose series, $e_0 = 0.852$. The individual test sheets of the source report (Ibsen and B{\o}dker, Data Report 9301, ``void ratio before test'') give the values in Table~\ref{tab:S1}: tests 9301.20--9301.29 belong to the medium-dense series and test 9301.32 to the dense series. The initial void ratios were corrected, and the void-ratio paths were rebuilt from the volumetric strain implied by the original files, $\varepsilon_{v,t} = (e_0^{\text{file}} - e_t^{\text{file}})/(1 + e_0^{\text{file}})$, as $e_t = e_0 - (1 + e_0)\,\varepsilon_{v,t}$. Figure~\ref{fig:S6} shows that, with the file values, specimens of nominally the same density and confining pressure reach clearly different peak deviatoric stresses, whereas with the corrected values the peak stresses order consistently with density.

\setcounter{table}{0}
\begin{table}[H]
\caption{Initial void ratios of Baskarp tests 9301.19--9301.32: values in the data files and values from the individual test sheets of the source report, and the role of each test in the split.}
\label{tab:S1}
\input{tables/S1_body}

\begin{figure}[H]
\centering
\includegraphics[width=\textwidth]{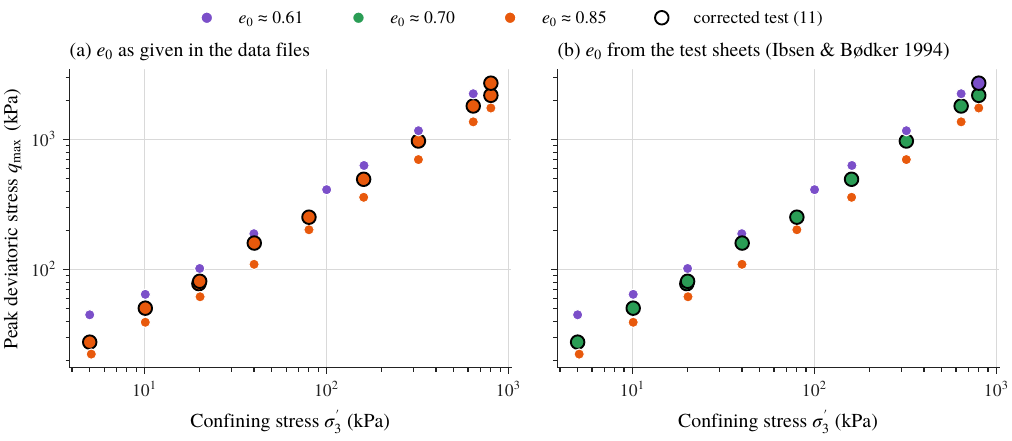}
\caption{Baskarp sand: peak deviatoric stress versus confining pressure, coloured by initial void ratio, (a) with the void ratios of the data files and (b) with the void ratios of the test sheets. Outlined markers: the 11 tests whose initial void ratio changed substantially.}
\label{fig:S6}
\end{figure}

\subsection*{S4. Window length and retained configurations}

\begin{figure}[H]
\centering
\includegraphics[width=\textwidth]{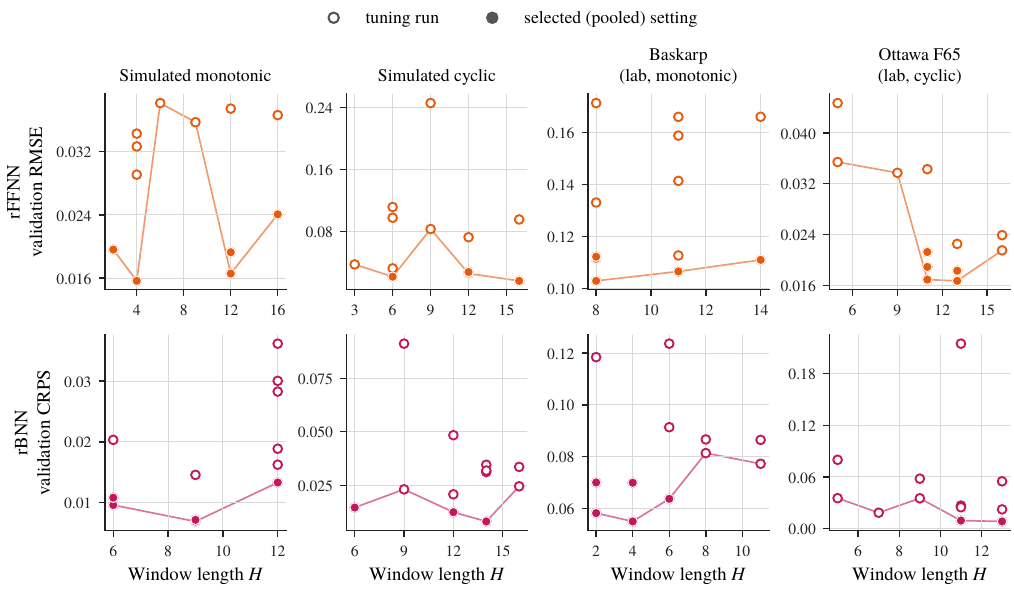}
\caption{Validation score of every tuning configuration of the rFFNN (RMSE, top) and the rBNN (CRPS, bottom) versus the window length $H$; the line connects the best score per $H$, and filled markers are the configurations retained in the pooled prediction. Each configuration differs in more than $H$ (\cref{sec:protocol} of the main text), so the scatter at a given $H$ reflects the other hyperparameters.}
\label{fig:S7}
\end{figure}

\setcounter{table}{1}
{\small
\input{tables/S2_pool}
}

\subsection*{S5. Training time}

\begin{figure}[H]
\centering
\includegraphics[width=\textwidth]{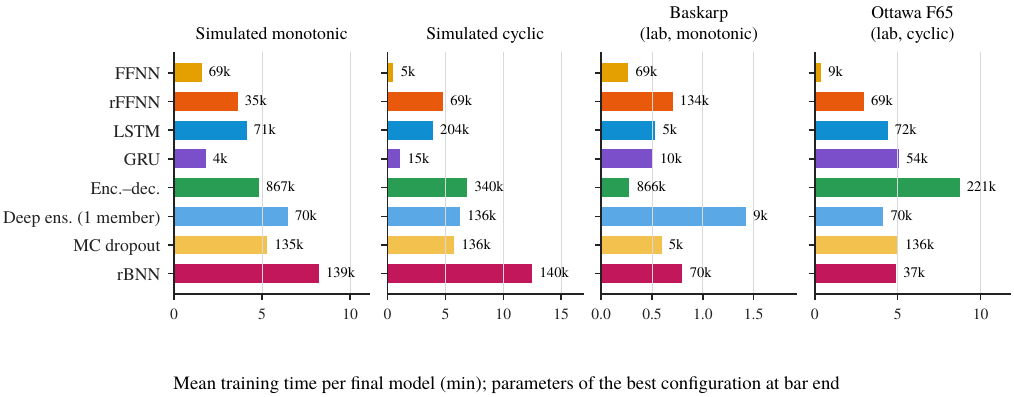}
\caption{Mean training time of one final model of each type (one NVIDIA RTX A6000 shared with other jobs; indicative values), with the number of trainable parameters of the best configuration at the end of each bar (for the rBNN, a mean and a scale per weight). Search times are given in \cref{tab:cost} of the main text.}
\label{fig:S8}
\end{figure}

}

\end{document}

%% file: figure1.tex
\begin{figure}[!ht]
\centering
\begin{subfigure}[b]{0.45\textwidth}  
\centering
\resizebox{0.9\textwidth}{!}{  
\begin{tikzpicture}    
[scale=1,
youngnode1/.style={rectangle, rounded corners, draw=pink!70, fill=pink!25, very thick, minimum size=45, drop shadow={shadow xshift=0.5mm, shadow yshift=-1mm, opacity=1}}, 
youngnode2/.style={rectangle, rounded corners, draw=yellow!70, fill=yellow!25, very thick, minimum size=45, drop shadow={shadow xshift=0.5mm, shadow yshift=-1mm, opacity=1}},
oldnode1/.style={rectangle, rounded corners, draw=green!, fill=green!25, very thick, minimum width=80, minimum height=120, drop shadow={shadow xshift=0.5mm, shadow yshift=-1mm, opacity=1}}, baseline
    ]
    
\node[oldnode1] (CucO) at (0,0) {FFNN};
\node[youngnode1] (SusO) at (-3,1.25) {$p'_t$};
\node[youngnode1] (InfO) [below=of SusO] {$\mathrm{d}\varepsilon_{t+1}$};
\node[youngnode2] (FnfO) [right=of CucO] {$p'_{t+1}$};

\draw[->,very thick] (SusO.east) -- (CucO.west |- SusO.east);
\draw[->,very thick] (InfO.east) -- (CucO.west |- InfO.east);
\draw[->,very thick] (CucO.east) -- (FnfO.west |- CucO.east);

\draw[->, very thick, draw = white ] 
    (FnfO.east) 
    -- ++(0.5, 0)  
    -- ++(0, -2.5) 
    -- ++(-8.60, 0) 
    -- ++(0, 3.75)  
    -- (SusO.west); 

\end{tikzpicture}}
\caption{FFNN}
\label{fig:fnn}
\end{subfigure}
\hfill
\begin{subfigure}[b]{0.45\textwidth}  
\centering
\resizebox{0.9\textwidth}{!}{  
\begin{tikzpicture}    
[scale=1,
youngnode1/.style={rectangle, rounded corners, draw=pink!70, fill=pink!25, very thick, minimum size=45, drop shadow={shadow xshift=0.5mm, shadow yshift=-1mm, opacity=1}}, 
youngnode2/.style={rectangle, rounded corners, draw=yellow!70, fill=yellow!25, very thick, minimum size=45, drop shadow={shadow xshift=0.5mm, shadow yshift=-1mm, opacity=1}},
oldnode1/.style={rectangle, rounded corners, draw=green!, fill=green!25, very thick, minimum width=80, minimum height=120, drop shadow={shadow xshift=0.5mm, shadow yshift=-1mm, opacity=1}}, baseline
    ]
    
\node[oldnode1] (CucO) at (0, 0) {\parbox{2cm}{\centering Recursive\\FFNN}};
\node[youngnode1] (SusO) at (-3,1.25) {$p'_t$};
\node[youngnode1] (InfO) [below=of SusO] {$\mathrm{d}\varepsilon_{t+1}$};
\node[youngnode2] (FnfO) [right=of CucO] {$p'_{t+1}$};

\draw[->,very thick] (SusO.east) -- (CucO.west |- SusO.east);
\draw[->,very thick] (InfO.east) -- (CucO.west |- InfO.east);
\draw[->,very thick] (CucO.east) -- (FnfO.west |- CucO.east);

\draw[->, very thick] 
    (FnfO.east) 
    -- ++(0.5, 0)  
    -- ++(0, -2.5) 
    -- ++(-8.60, 0) 
    -- ++(0, 3.75)  
    -- (SusO.west); 

\end{tikzpicture}}
\caption{Recursive-FFNN}
\label{fig:fbnn}
\end{subfigure}
\hfill

\caption{ \textbf{Comparison of FFNN and rFFNN training approach}: (a) A single-step FFNN utilizes the measured current stress state and strain increment to predict the subsequent stress state. (b) The rFFNN incorporates the predicted stress state from the current step as feedback to predict the next stress state, recursively incorporating prior predictions as inputs to capture sequential and path-dependent stress-strain behavior.}
\label{fig:fnn_fbnn}
\end{figure}

%% file: tables/hyper.tex
{
\begin{table}[!ht]

\scriptsize
\caption{\textbf{Best configuration of every model} (rank 1 of 12 random configurations on the validation set).}
\label{tab:hyper}
\begin{tabularx}{\textwidth}{lCCCCCCCCCCCC}
\toprule
 & \multicolumn{3}{c}{Sim. monotonic} & \multicolumn{3}{c}{Sim. cyclic} & \multicolumn{3}{c}{Baskarp} & \multicolumn{3}{c}{Ottawa F65} \\
\cmidrule(lr){2-4}\cmidrule(lr){5-7}\cmidrule(lr){8-10}\cmidrule(lr){11-13}
Model & $H$ & size & params & $H$ & size & params & $H$ & size & params & $H$ & size & params \\
\midrule
FFNN & 1 & 2$\times$256 & 69k & 1 & 2$\times$64 & 5k & 1 & 2$\times$256 & 69k & 1 & 3$\times$64 & 9k \\
rFFNN & 4 & 3$\times$128 & 35k & 16 & 2$\times$256 & 69k & 8 & 3$\times$256 & 134k & 13 & 2$\times$256 & 69k \\
LSTM & -- & 1$\times$128 & 71k & -- & 2$\times$128 & 204k & -- & 1$\times$32 & 5k & -- & 1$\times$128 & 72k \\
GRU & -- & 1$\times$32 & 4k & -- & 1$\times$64 & 15k & -- & 2$\times$32 & 10k & -- & 1$\times$128 & 54k \\
Encoder--decoder & -- & 2$\times$128 & 867k & -- & 1$\times$128 & 340k & -- & 2$\times$128 & 866k & -- & 2$\times$64 & 221k \\
Deep ensemble (member) & 16 & 2$\times$256 & 70k & 6 & 3$\times$256 & 136k & 8 & 3$\times$64 & 9k & 11 & 2$\times$256 & 70k \\
MC dropout & 12 & 3$\times$256 & 135k & 6 & 3$\times$256 & 136k & 6 & 2$\times$64 & 5k & 7 & 3$\times$256 & 136k \\
rBNN & 9 & 2$\times$256 & 139k & 14 & 2$\times$256 & 140k & 4 & 3$\times$128 & 70k & 13 & 2$\times$128 & 37k \\
\bottomrule
\end{tabularx}
\par\vspace{2pt}{\footnotesize $H$: window length (number of recursive predictions per training window; $H = 1$ is one-step training). Size: hidden layers $\times$ units (recurrent layers for LSTM and GRU; encoder and decoder layers for the encoder--decoder). Params: trainable parameters; for the rBNN, a mean and a scale per weight. All retained configurations are listed in Supplementary Table S2.}
\end{table}}

%% file: tables/accuracy.tex
{
\begin{table}[!ht]

\footnotesize
\caption{\textbf{Test-set accuracy of all models.} Every test curve is predicted recursively from its initial state; each row is the pooled prediction of the retained configurations of that model (\cref{sec:protocol}). $R^2$ is computed on the $\sigma'_3$-normalized stresses. Best value per column and dataset in bold.}
\label{tab:accuracy}
\begin{tabularx}{\textwidth}{lCCCCCCCCCC}
\toprule
 & \multicolumn{3}{c}{RMSE} & \multicolumn{3}{c}{MAE} & \multicolumn{4}{c}{$R^2$} \\
\cmidrule(lr){2-4}\cmidrule(lr){5-7}\cmidrule(lr){8-11}
Model & $p'$ (kPa) & $q$ (kPa) & 3rd & $p'$ (kPa) & $q$ (kPa) & 3rd & $p'$ & $q$ & 3rd & mean \\
\midrule
\multicolumn{11}{l}{\textit{Simulated monotonic} (third state variable: void ratio $e$, errors in units of $10^{-6}$)}\\
\quad FFNN & 7.5 & 22.4 & 8.4 & 4.5 & 13.5 & 6.1 & 0.979 & 0.979 & 0.997 & 0.985 \\
\quad rFFNN & 4.2 & 12.5 & 8.8 & 2.3 & 6.8 & 6.9 & 0.984 & 0.984 & 0.997 & 0.988 \\
\quad LSTM & 4.6 & 13.6 & 5.8 & 3.0 & 9.0 & 4.6 & \textbf{0.999} & \textbf{0.999} & 0.999 & \textbf{0.999} \\
\quad GRU & \textbf{2.4} & \textbf{7.1} & 3.4 & 1.5 & 4.3 & 2.8 & 0.991 & 0.991 & \textbf{1.000} & 0.994 \\
\quad Encoder--decoder & 2.6 & 7.7 & 4.6 & 1.5 & 4.4 & 3.4 & 0.987 & 0.987 & 0.999 & 0.991 \\
\quad Deep ensemble & 2.5 & 7.7 & \textbf{2.2} & \textbf{1.1} & \textbf{3.3} & \textbf{1.7} & 0.996 & 0.996 & \textbf{1.000} & 0.998 \\
\quad MC dropout & 7.2 & 21.6 & 6.3 & 4.0 & 12.0 & 5.0 & 0.991 & 0.991 & 0.998 & 0.993 \\
\quad rBNN & 2.7 & 8.0 & 3.4 & 1.4 & 4.3 & 2.1 & 0.988 & 0.988 & \textbf{1.000} & 0.992 \\
\addlinespace[2pt]
\multicolumn{11}{l}{\textit{Simulated cyclic} (third state variable: volumetric strain $\varepsilon_v$, errors in units of $10^{-4}$)}\\
\quad FFNN & 33.2 & 84.5 & 22.2 & 18.9 & 51.6 & 13.5 & 0.925 & 0.801 & 0.950 & 0.892 \\
\quad rFFNN & 9.2 & 13.8 & 4.3 & 3.5 & 5.8 & 2.9 & 0.994 & 0.995 & 0.998 & 0.996 \\
\quad LSTM & \textbf{6.7} & \textbf{9.6} & \textbf{1.7} & \textbf{1.9} & \textbf{2.7} & \textbf{1.0} & \textbf{0.997} & \textbf{0.998} & \textbf{1.000} & \textbf{0.998} \\
\quad GRU & 7.4 & 10.6 & 2.8 & 2.6 & 4.1 & 2.0 & 0.996 & \textbf{0.998} & 0.999 & \textbf{0.998} \\
\quad Encoder--decoder & 10.9 & 14.1 & 4.5 & 3.3 & 4.4 & 2.4 & 0.992 & 0.996 & 0.998 & 0.995 \\
\quad Deep ensemble & 8.9 & 12.1 & 3.9 & 4.1 & 5.4 & 2.5 & 0.994 & 0.996 & 0.998 & 0.996 \\
\quad MC dropout & 18.7 & 28.0 & 10.5 & 7.6 & 12.7 & 6.1 & 0.976 & 0.971 & 0.989 & 0.979 \\
\quad rBNN & 10.5 & 14.5 & 4.0 & 3.9 & 8.2 & 2.7 & 0.993 & 0.993 & 0.998 & 0.995 \\
\addlinespace[2pt]
\multicolumn{11}{l}{\textit{Baskarp sand} (third state variable: void ratio $e$, errors in units of $10^{-3}$)}\\
\quad FFNN & 37.0 & 97.4 & 9.7 & 19.8 & 58.4 & 6.8 & 0.850 & 0.874 & 0.988 & 0.904 \\
\quad rFFNN & 36.6 & 103.0 & 8.3 & 20.5 & 57.4 & 6.3 & 0.927 & 0.921 & 0.991 & 0.946 \\
\quad LSTM & 25.2 & 78.7 & 8.6 & 14.2 & 43.2 & 6.5 & \textbf{0.934} & \textbf{0.944} & 0.991 & \textbf{0.956} \\
\quad GRU & \textbf{17.0} & \textbf{57.3} & 7.1 & \textbf{9.9} & \textbf{32.5} & 5.6 & 0.921 & 0.924 & \textbf{0.994} & 0.946 \\
\quad Encoder--decoder & 38.3 & 94.9 & 11.6 & 22.3 & 56.8 & 8.4 & 0.843 & 0.884 & 0.983 & 0.904 \\
\quad Deep ensemble & 31.6 & 105.2 & 23.4 & 18.2 & 62.4 & 20.6 & 0.915 & 0.914 & 0.932 & 0.920 \\
\quad MC dropout & 28.4 & 96.3 & \textbf{6.8} & 15.8 & 57.6 & \textbf{4.6} & 0.919 & 0.917 & \textbf{0.994} & 0.944 \\
\quad rBNN & 24.8 & 63.3 & 11.0 & 15.1 & 36.3 & 8.7 & 0.920 & 0.934 & 0.985 & 0.946 \\
\addlinespace[2pt]
\multicolumn{11}{l}{\textit{Ottawa F65 sand} (third state variable: pore pressure ratio $r_u$, errors in units of $10^{-2}$)}\\
\quad FFNN & 52.5 & 61.3 & 24.9 & 35.5 & 38.9 & 17.9 & $-7.4$ & $-10.9$ & $-1.7$ & $-6.6$ \\
\quad rFFNN & 3.6 & 3.5 & 3.6 & 2.8 & 2.6 & 2.7 & 0.962 & 0.961 & 0.945 & 0.956 \\
\quad LSTM & 2.3 & \textbf{2.3} & \textbf{2.1} & \textbf{1.2} & 1.6 & \textbf{1.2} & 0.984 & \textbf{0.983} & \textbf{0.982} & \textbf{0.983} \\
\quad GRU & \textbf{2.2} & 2.5 & 2.4 & 1.5 & 1.6 & 1.6 & \textbf{0.985} & 0.980 & 0.976 & 0.980 \\
\quad Encoder--decoder & \textbf{2.2} & 2.4 & 2.2 & \textbf{1.2} & \textbf{1.5} & 1.3 & \textbf{0.985} & 0.982 & 0.980 & 0.982 \\
\quad Deep ensemble & 2.7 & 4.4 & 2.8 & 2.0 & 2.5 & 2.1 & 0.978 & 0.938 & 0.965 & 0.961 \\
\quad MC dropout & 19.4 & 9.0 & 18.4 & 13.9 & 5.9 & 13.0 & $-0.145$ & 0.743 & $-0.455$ & 0.048 \\
\quad rBNN & 3.0 & 4.5 & 3.4 & 1.8 & 2.4 & 2.0 & 0.973 & 0.935 & 0.951 & 0.953 \\
\bottomrule
\end{tabularx}
\end{table}}

%% file: tables/uncertainty.tex
{
\begin{table}[!ht]

\footnotesize
\caption{\textbf{Quality of the predictive intervals} of the three probabilistic models on the test sets.}
\label{tab:uncertainty}
\begin{tabularx}{\textwidth}{lCCCCCC}
\toprule
 & Accuracy & \multicolumn{3}{c}{Reliability of the 95\% interval} & Sharpness & Overall \\
\cmidrule(lr){2-2}\cmidrule(lr){3-5}\cmidrule(lr){6-6}\cmidrule(lr){7-7}
Model & mean $R^2$ & PICP$_{95}$ & reliable specimens$^{a}$ & worst specimen$^{b}$ & width$^{c}$ & interval score$^{d}$ \\
\midrule
\multicolumn{7}{l}{\textit{Simulated monotonic}}\\
\quad rBNN & 0.99 & 0.99 & \textbf{6}/6 & 0.91 & \textbf{0.09} & 0.09 \\
\quad Deep ensemble & \textbf{1.00} & 0.96 & 4$^{\dagger}$/6 & 0.76 & 0.03 & \textbf{0.04} \\
\quad MC dropout & 0.99 & 1.00 & \textbf{6}/6 & \textbf{1.00} & 0.12 & 0.12 \\
\addlinespace[2pt]
\multicolumn{7}{l}{\textit{Simulated cyclic}}\\
\quad rBNN & 0.99 & 0.99 & 5/6 & 0.86 & \textbf{0.06} & 0.08 \\
\quad Deep ensemble & \textbf{1.00} & 0.95 & 4$^{\dagger}$/6 & 0.75 & 0.04 & \textbf{0.07} \\
\quad MC dropout & 0.98 & 1.00 & \textbf{6}/6 & \textbf{0.97} & 0.22 & 0.23 \\
\addlinespace[2pt]
\multicolumn{7}{l}{\textit{Baskarp sand}}\\
\quad rBNN & \textbf{0.95} & 0.97 & \textbf{5}/6 & 0.65 & 0.26 & 0.31 \\
\quad Deep ensemble & 0.92 & 0.87 & 2$^{\dagger}$/6 & 0.22 & 0.22 & 0.51 \\
\quad MC dropout & 0.94 & 0.97 & \textbf{5}/6 & \textbf{0.70} & \textbf{0.25} & \textbf{0.29} \\
\addlinespace[2pt]
\multicolumn{7}{l}{\textit{Ottawa F65 sand}}\\
\quad rBNN & 0.95 & 0.99 & \textbf{3}/3 & 0.97 & \textbf{0.11} & \textbf{0.11} \\
\quad Deep ensemble & \textbf{0.96} & 1.00 & \textbf{3}/3 & \textbf{1.00} & 0.16 & 0.16 \\
\quad MC dropout & 0.05$^{\dagger}$ & 1.00 & \textbf{3}/3 & \textbf{1.00} & 1.03$^{\dagger}$ & 0.94 \\
\bottomrule
\end{tabularx}
\par\vspace{2pt}{\footnotesize $^{a}$\,Test specimens whose 95\% interval contains at least 90\% of the measured points for each of the three state variables. $^{b}$\,Lowest coverage of any test specimen (worst of the three state variables). $^{c}$\,Median over test specimens of the mean interval width divided by the range of the test data. $^{d}$\,Interval score of the 95\% interval \cite{gneiting2007strictly}, divided by the data range; lower is better. $^{\dagger}$\,Fails a requirement: mean $R^2 < 0.90$, fewer than 80\% of specimens reliable, or interval wider than half the data range. Bold: best value (width: narrowest among models without $^{\dagger}$).}
\end{table}}

%% file: tables/ablation.tex
{
\begin{table}[!ht]

\scriptsize
\caption{\textbf{Ablation study}: one ingredient changed at a time. Mean test $R^2$ and the lowest 95\% coverage (PICP) of the three state variables; point models have no coverage. Every variant is tuned, retrained and pooled in the same way. A5 is the proposed rBNN.}
\label{tab:ablation}
\begin{tabularx}{\textwidth}{l>{\raggedright\arraybackslash}p{5.0cm}CCCCCCCC}
\toprule
 & & \multicolumn{2}{c}{Sim. monotonic} & \multicolumn{2}{c}{Sim. cyclic} & \multicolumn{2}{c}{Baskarp} & \multicolumn{2}{c}{Ottawa F65} \\
\cmidrule(lr){3-4}\cmidrule(lr){5-6}\cmidrule(lr){7-8}\cmidrule(lr){9-10}
 & Variant & $R^2$ & PICP & $R^2$ & PICP & $R^2$ & PICP & $R^2$ & PICP \\
\midrule
A1 & FFNN: one-step training, point prediction & 0.985 & -- & 0.892 & -- & 0.904 & -- & $-6.6$ & -- \\
A2 & + recursive window training (rFFNN) & 0.988 & -- & 0.996 & -- & 0.946 & -- & 0.956 & -- \\
A3 & + variance output, pooled (deep ensemble) & 0.998 & 0.93 & 0.996 & 0.93 & 0.920 & 0.80 & 0.961 & 1.00 \\
A4 & Bayesian weights, one-step training ($H = 1$) & 0.885 & 1.00 & 0.385 & 0.99 & 0.931 & 0.99 & $-1.1$ & 0.93 \\
\textbf{A5} & Bayesian weights + recursive training (rBNN) & 0.992 & 0.98 & 0.995 & 0.98 & 0.946 & 0.95 & 0.953 & 0.98 \\
A6 & rBNN, R\'enyi divergence, $\alpha = 0.5$ & 0.991 & 1.00 & 0.991 & 1.00 & 0.941 & 0.94 & 0.950 & 0.95 \\
A7 & rBNN, R\'enyi divergence, $\alpha = 2$ & 0.993 & 0.98 & 0.997 & 0.97 & 0.947 & 0.94 & 0.949 & 0.99 \\
A8 & rBNN, KL term down-weighted ($10^{-4}\lambda_{\text{KL}}$) & 0.994 & 0.99 & 0.995 & 0.95 & 0.933 & 0.92 & 0.907 & 0.93 \\
A9 & rBNN, new weight sample at every step at prediction & 0.992 & 0.93 & 0.995 & 0.95 & 0.946 & 0.94 & 0.948 & 0.97 \\
A10 & rBNN, sampled (not mean) state fed back & 0.991 & 0.99 & 0.994 & 0.99 & 0.946 & 0.97 & 0.944 & 0.99 \\
\bottomrule
\end{tabularx}
\end{table}}

%% file: tables/cost.tex
{
\begin{table}[!ht]

\scriptsize
\caption{\textbf{Training time} of one final model (mean over the retrained configurations) and of the complete hyperparameter search.}
\label{tab:cost}
\begin{tabularx}{\textwidth}{lCCCCCCCCC}
\toprule
 & \multicolumn{4}{c}{One final model (min)} & \multicolumn{5}{c}{Complete search, 12 configurations (min)} \\
\cmidrule(lr){2-5}\cmidrule(lr){6-10}
Model & Sim. mono. & Sim. cyc. & Baskarp & Ottawa & Sim. mono. & Sim. cyc. & Baskarp & Ottawa & Total \\
\midrule
FFNN & 1.6 & 0.5 & 0.3 & 0.4 & 11 & 3 & 6 & 4 & 24 \\
rFFNN & 3.6 & 4.8 & 0.7 & 3.0 & 40 & 42 & 18 & 43 & 144 \\
LSTM & 4.1 & 3.9 & 0.5 & 4.4 & 27 & 60 & 8 & 35 & 130 \\
GRU & 1.8 & 1.1 & 0.5 & 5.1 & 15 & \textbf{223} & 6 & 58 & 302 \\
Encoder--decoder & 4.8 & 6.9 & 0.3 & 8.7 & 41 & 74 & 9 & 125 & 248 \\
Deep ensemble (member) & 6.5 & 6.3 & 1.4 & 4.1 & 73 & 57 & 23 & 63 & 216 \\
MC dropout & 5.3 & 5.8 & 0.6 & 5.0 & 57 & 72 & 19 & 89 & 237 \\
rBNN & 8.2 & 12.5 & 0.8 & 4.9 & \textbf{97} & 145 & \textbf{28} & \textbf{130} & \textbf{401} \\
\bottomrule
\end{tabularx}
\par\vspace{2pt}{\footnotesize Wall-clock training time on one NVIDIA RTX A6000 shared with other jobs; values are indicative and best compared within a column. Bold: largest search time per dataset.}
\end{table}}

%% file: figure2.tex
\begin{figure}[H]
\hspace*{-1cm}
\centering
\begin{tikzpicture}[every node/.style={anchor=center}, scale = 0.7]

\matrix (m1) [matrix of math nodes,left delimiter={(},right delimiter={)},
nodes in empty cells, column sep=0.01cm, row sep=0.01cm] {
\textcolor{red}{\ve{s}_0^{(1)}} & \textcolor{red}{\ve{u}_1^{(1)}}  & \textcolor{red}{\ve{\theta}^{(1)}} & \hdots  & \textcolor{red}{\ve{s}_{N-1}^{(1)}} & \textcolor{red}{\ve{u}_N^{(1)}}  & \textcolor{red}{\ve{\theta}^{(1)}} \\
\textcolor{blue}{\ve{s}_0^{(2)}} & \textcolor{blue}{\ve{u}_1^{(2)}}  & \textcolor{blue}{\ve{\theta}^{(2)}} & \hdots  & \textcolor{blue}{\ve{s}_{N-1}^{(2)}} & \textcolor{blue}{\ve{u}_N^{(2)}}  & \textcolor{blue}{\ve{\theta}^{(2)}} \\
 & \vdots  &  \\
\textcolor{purple}{\ve{s}_0^{(M)}} & \textcolor{purple}{\ve{u}_1^{(M)}}  & \textcolor{purple}{\ve{\theta}^{(M)}} & \hdots  & \textcolor{purple}{\ve{s}_{N-1}^{(M)}} & \textcolor{purple}{\ve{u}_N^{(M)}}  & \textcolor{purple}{\ve{\theta}^{(M)}} \\
};

\matrix (m2) [matrix of math nodes,left delimiter={(},right delimiter={)},
nodes in empty cells, column sep=0.01cm, row sep=0.01cm, right=1cm of m1] {
\textcolor{red}{\ve{s}_N^{(1)}} \\
\textcolor{blue}{\ve{s}_N^{(1)}} \\
\vdots \\
\textcolor{purple}{\ve{s}_N^{(1)}} \\
};

\draw[->, thick] ($(m1.east)+(0.5,0)$) -- ($(m2.west)+(-0.5,0)$);

\end{tikzpicture}
\caption{Illustration of the data structure used for Recurrent Neural Networks.}
\label{fig:sample_ffnn_single}
\end{figure}

%% file: tables/S3_single.tex
\begin{table}[!ht]\centering\small
\caption{Coverage of the 95\% interval of single networks (before pooling): median and range over the five retrained configurations of the PICP$_{95}$ averaged over the three state variables, compared with the pooled prediction used in the main text.}\label{tab:S3}
\begin{tabular}{llccc}
\toprule
Dataset & Model & single network, median & single network, range & pooled \\
\midrule
Simulated monotonic & rBNN & 0.90 & 0.68--1.00 & 0.99 \\
 & Deep ensemble & 0.51 & 0.01--0.98 & 0.96 \\
 & MC dropout & 0.94 & 0.70--1.00 & 1.00 \\
\addlinespace[2pt]
Simulated cyclic & rBNN & 0.97 & 0.95--0.99 & 0.99 \\
 & Deep ensemble & 0.67 & 0.55--0.77 & 0.95 \\
 & MC dropout & 0.97 & 0.94--1.00 & 1.00 \\
\addlinespace[2pt]
Baskarp sand & rBNN & 0.90 & 0.82--0.96 & 0.97 \\
 & Deep ensemble & 0.53 & 0.26--0.65 & 0.87 \\
 & MC dropout & 0.94 & 0.87--0.99 & 0.97 \\
\addlinespace[2pt]
Ottawa F65 sand & rBNN & 0.97 & 0.95--1.00 & 0.99 \\
 & Deep ensemble & 0.94 & 0.74--0.97 & 1.00 \\
 & MC dropout & 1.00 & 0.99--1.00 & 1.00 \\
\bottomrule
\end{tabular}
\end{table}

%% file: tables/S1_body.tex
\centering\small
\begin{tabular}{lcccc}
\toprule
Test & $\sigma'_3$ (kPa) & $e_0$ in data file & $e_0$ from test sheet & Role \\
\midrule
9301.19 & 5.1 & 0.852 & 0.852 & test\\
9301.20 & 80.1 & 0.852 & 0.705 & training\\
9301.21 & 160.0 & 0.852 & 0.695 & training\\
9301.22 & 40.1 & 0.852 & 0.703 & training\\
9301.23 & 20.1 & 0.852 & 0.702 & training\\
9301.24 & 10.1 & 0.852 & 0.696 & validation\\
9301.25 & 19.9 & 0.852 & 0.696 & training\\
9301.26 & 5.0 & 0.852 & 0.704 & test\\
9301.27 & 320.0 & 0.852 & 0.698 & training\\
9301.28 & 640.1 & 0.852 & 0.698 & validation\\
9301.29 & 800.2 & 0.852 & 0.699 & test\\
9301.30 & 640.2 & 0.852 & 0.856 & validation\\
9301.31 & 800.0 & 0.852 & 0.846 & test\\
9301.32 & 800.2 & 0.852 & 0.614 & test\\
\bottomrule
\end{tabular}
\end{table}

%% file: tables/S2_pool.tex
\begin{longtable}{p{2.6cm}p{2.6cm}cp{8.2cm}}
\caption{Configurations retained in the pooled prediction of every model (of the five best of 12, those whose validation score is at most twice the best), in order of validation score. $H$: window length in the paper's notation; layers $\times$ units.}\label{tab:S2}\\
\toprule
Dataset & Model & $n$ & Configurations \\
\midrule
\endfirsthead
\toprule
Dataset & Model & $n$ & Configurations \\
\midrule
\endhead
Simulated monotonic & FFNN & 4 & $H$=1, 2$\times$256; $H$=1, 3$\times$128; $H$=1, 3$\times$128; $H$=1, 2$\times$256 \\
 & rFFNN & 5 & $H$=4, 3$\times$128; $H$=12, 3$\times$128; $H$=12, 3$\times$64; $H$=2, 3$\times$128; $H$=16, 2$\times$64 \\
 & LSTM & 1 & 1$\times$128 \\
 & GRU & 5 & 1$\times$32; 1$\times$32; 2$\times$64; 1$\times$64; 1$\times$64 \\
 & Encoder--decoder & 3 & 2$\times$128; 1$\times$32; 2$\times$64 \\
 & Deep ensemble & 3 & $H$=16, 2$\times$256; $H$=16, 2$\times$256; $H$=9, 2$\times$256 \\
 & MC dropout & 4 & $H$=12, 3$\times$256; $H$=12, 3$\times$128; $H$=9, 3$\times$256; $H$=16, 3$\times$128 \\
 & rBNN & 5 & $H$=9, 2$\times$256; $H$=9, 2$\times$128; $H$=6, 3$\times$256; $H$=6, 2$\times$128; $H$=12, 2$\times$64 \\
Simulated cyclic & FFNN & 3 & $H$=1, 2$\times$64; $H$=1, 3$\times$256; $H$=1, 2$\times$128 \\
 & rFFNN & 4 & $H$=16, 2$\times$256; $H$=6, 3$\times$128; $H$=12, 2$\times$128; $H$=12, 3$\times$128 \\
 & LSTM & 5 & 2$\times$128; 1$\times$64; 2$\times$128; 1$\times$128; 2$\times$64 \\
 & GRU & 5 & 1$\times$64; 1$\times$128; 2$\times$32; 1$\times$32; 1$\times$32 \\
 & Encoder--decoder & 4 & 1$\times$128; 2$\times$128; 2$\times$128; 2$\times$32 \\
 & Deep ensemble & 4 & $H$=6, 3$\times$256; $H$=16, 3$\times$256; $H$=16, 2$\times$128; $H$=14, 2$\times$64 \\
 & MC dropout & 5 & $H$=6, 3$\times$256; $H$=9, 3$\times$64; $H$=16, 3$\times$64; $H$=6, 3$\times$64; $H$=12, 2$\times$128 \\
 & rBNN & 3 & $H$=14, 2$\times$256; $H$=12, 3$\times$64; $H$=6, 2$\times$256 \\
Baskarp sand & FFNN & 5 & $H$=1, 2$\times$256; $H$=1, 3$\times$256; $H$=1, 3$\times$128; $H$=1, 3$\times$256; $H$=1, 2$\times$256 \\
 & rFFNN & 5 & $H$=8, 3$\times$256; $H$=11, 2$\times$256; $H$=14, 2$\times$128; $H$=8, 2$\times$256; $H$=8, 3$\times$256 \\
 & LSTM & 5 & 1$\times$32; 1$\times$32; 1$\times$128; 2$\times$64; 2$\times$32 \\
 & GRU & 5 & 2$\times$32; 1$\times$128; 2$\times$64; 1$\times$32; 1$\times$128 \\
 & Encoder--decoder & 5 & 2$\times$128; 2$\times$128; 1$\times$128; 2$\times$64; 2$\times$64 \\
 & Deep ensemble & 5 & $H$=8, 3$\times$64; $H$=4, 3$\times$64; $H$=8, 3$\times$128; $H$=4, 2$\times$64; $H$=4, 3$\times$128 \\
 & MC dropout & 5 & $H$=6, 2$\times$64; $H$=2, 3$\times$256; $H$=14, 3$\times$64; $H$=14, 2$\times$128; $H$=6, 2$\times$64 \\
 & rBNN & 5 & $H$=4, 3$\times$128; $H$=2, 3$\times$128; $H$=6, 2$\times$256; $H$=4, 3$\times$256; $H$=2, 2$\times$64 \\
Ottawa F65 sand & FFNN & 5 & $H$=1, 3$\times$64; $H$=1, 2$\times$128; $H$=1, 3$\times$128; $H$=1, 3$\times$64; $H$=1, 3$\times$128 \\
 & rFFNN & 5 & $H$=13, 2$\times$256; $H$=11, 2$\times$128; $H$=13, 2$\times$128; $H$=11, 2$\times$256; $H$=11, 2$\times$64 \\
 & LSTM & 5 & 1$\times$128; 1$\times$32; 2$\times$32; 1$\times$32; 2$\times$64 \\
 & GRU & 5 & 1$\times$128; 2$\times$128; 2$\times$128; 2$\times$128; 2$\times$64 \\
 & Encoder--decoder & 5 & 2$\times$64; 2$\times$32; 1$\times$64; 1$\times$32; 1$\times$32 \\
 & Deep ensemble & 5 & $H$=11, 2$\times$256; $H$=13, 2$\times$64; $H$=11, 3$\times$128; $H$=16, 2$\times$64; $H$=16, 3$\times$256 \\
 & MC dropout & 4 & $H$=7, 3$\times$256; $H$=9, 3$\times$256; $H$=16, 3$\times$256; $H$=13, 2$\times$256 \\
 & rBNN & 2 & $H$=13, 2$\times$128; $H$=11, 2$\times$128 \\
\bottomrule
\end{longtable}